%% file: PFNM_arxiv.tex
\documentclass{article}

\usepackage{microtype}
\usepackage{booktabs} 
\usepackage{flushend}

\usepackage[accepted]{icml2019}

\input{math_commands.tex}

\usepackage{xr}

\usepackage{hyperref}
\usepackage{url}
\usepackage{comment}
\usepackage{amsmath, amssymb,amsthm, color}
\usepackage{amsthm} 
\usepackage{algorithm,algorithmic}
\usepackage{graphicx}
\usepackage{subcaption}
\usepackage{wrapfig}
\usepackage{mathtools}
\usepackage{cleveref}
\newcommand{\Tp}{\ensuremath{\mathcal{T}}}
\DeclareMathOperator*{\card}{card}

\newtheorem{definition}{Definition}

\newtheorem{remark}{Remark}

\newtheorem{proposition}{Proposition}

\icmltitlerunning{Bayesian Nonparametric Federated Learning of Neural Networks}

\begin{document}

\twocolumn[
\icmltitle{Bayesian Nonparametric Federated Learning of Neural Networks}



\icmlsetsymbol{equal}{*}

\begin{icmlauthorlist}
\icmlauthor{Mikhail Yurochkin}{to,To}
\icmlauthor{Mayank Agarwal}{to,To}
\icmlauthor{Soumya Ghosh}{to,To,too}
\icmlauthor{Kristjan Greenewald}{to,To}
\icmlauthor{Trong Nghia Hoang}{to,To}
\icmlauthor{Yasaman Khazaeni}{to,To}
\end{icmlauthorlist}

\icmlaffiliation{to}{IBM Research, Cambridge}
\icmlaffiliation{To}{MIT-IBM Watson AI Lab}
\icmlaffiliation{too}{Center for Computational Health}

\icmlcorrespondingauthor{Mikhail Yurochkin}{mikhail.yurochkin@ibm.com}

\icmlkeywords{Federated Learning, BNP, IBP}

\vskip 0.3in
]



\printAffiliationsAndNotice{}  

\input{abstract.tex}
\input{intro.tex}
\input{background.tex}

\section{Probabilistic Federated Neural Matching}
\label{sec:matching}
We now describe how the Bayesian nonparametric machinery can be applied to the problem of federated learning with neural networks. Our goal will be to identify subsets of neurons in each of the $J$ local models that match neurons in other local models. We will then appropriately combine the matched neurons to form a global model. 

Our approach to federated learning builds upon the following basic problem. Suppose we have trained $J$ Multilayer Perceptrons (MLPs) with one hidden layer each. For the $j$th MLP $j = 1,\dots, J$, let $V_j^{(0)} \in \mathbb{R}^{D\times L_j}$ and $\tilde{v}_j^{(0)} \in \mathbb{R}^{L_j}$ be the weights and biases of the hidden layer; $V_j^{(1)} \in \mathbb{R}^{L_j \times K}$ and $\tilde{v}_j^{(1)} \in \mathbb{R}^{K}$ be weights and biases of the softmax layer; $D$ be the data dimension, $L_j$ the number of neurons on the hidden layer; and $K$ the number of classes. We consider a simple architecture: $f_j(x) = \text{softmax}(\sigma(xV_j^{(0)} + \tilde{v}_j^{(0)})V_j^{(1)} + \tilde{v}_j^{(1)})$ where $\sigma(\cdot)$ is some nonlinearity (sigmoid, ReLU, etc.). Given the collection of weights and biases $\{V_j^{(0)},\tilde{v}_j^{(0)},V_j^{(1)},\tilde{v}_j^{(1)}\}_{j=1}^J$ we want to learn a global neural network with weights and biases $\Theta^{(0)}\in \mathbb{R}^{D\times L},\tilde{\theta}^{(0)} \in \mathbb{R}^L, \Theta^{(1)}\in \mathbb{R}^{L\times K},\tilde{\theta}^{(1)} \in \mathbb{R}^K$, where $L \ll \sum_{j=1}^J L_j$ is an unknown number of hidden units of the global network to be inferred. 

Our first observation is that ordering of neurons of the hidden layer of an MLP is permutation invariant. Consider any permutation $\tau(1,\ldots,L_j)$ of the $j$-th MLP -- reordering columns of $V_j^{(0)}$, biases $\tilde{v}_j^{(0)}$ and rows of $V_j^{(1)}$ according to $\tau(1,\ldots,L_j)$ will not affect the outputs $f_j(x)$ for any value of $x$. Therefore, instead of treating weights as matrices and biases as vectors we view them as unordered collections of vectors $V_j^{(0)} = \{v_{jl}^{(0)}\in \mathbb{R}^D\}_{l=1}^{L_j}$, $V_j^{(1)} = \{v_{jl}^{(1)}\in \mathbb{R}^{L_j}\}_{l=1}^{K}$ and scalars $\tilde{v}_j^{(0)} = \{\tilde{v}_{jl}^{(0)} \in \mathbb{R}\}_{l=1}^{L_j}$ correspondingly.

Hidden layers in neural networks are commonly viewed as feature extractors. This perspective can be justified by the fact that the last layer of a neural network classifier simply performs a softmax regression. Since neural networks often outperform basic softmax regression, they must be learning high quality feature representations of the raw input data. Mathematically, in our  setup, every hidden neuron of the $j$-th MLP represents a new feature $\tilde{x}_l(v^{(0)}_{jl},\tilde{v}_{jl}^{(0)}) = \sigma(\langle x, v^{(0)}_{jl}\rangle + \tilde{v}_{jl}^{(0)})$. Our second observation is that each ($v^{(0)}_{jl},\tilde{v}_{jl}^{(0)})$ parameterizes the corresponding neuron's feature extractor. Since, the $J$ MLPs are trained on the same general type of data (not necessarily homogeneous), we assume that they share at least some feature extractors that serve the same purpose. However, due to the permutation invariance issue discussed previously, a feature extractor indexed by $l$ from the $j$-th MLP is unlikely to correspond to a feature extractor with the same index from a different MLP. In order to construct a set of global feature extractors (neurons) $\{\theta_i^{(0)} \in \mathbb{R}^D, \tilde{\theta}_i^{(0)} \in \mathbb{R}\}_{i=1}^L$ we must model the process of grouping and combining feature extractors of collection of MLPs.

\subsection{Single Layer Neural Matching}
\label{single-layer}
We now present the key building block of our framework, a Beta Bernoulli Process \citep{thibaux2007hierarchical} based model of MLP weight parameters. Our model assumes the following generative process. First, draw a collection of global atoms (hidden layer neurons) from a Beta process prior with a base measure $H$
and mass parameter $\gamma_0$,  $Q = \sum_i \rq_i \delta_{\rvtheta_i}$. In our experiments we choose $H = \mathcal{N}(\vmu_0, \mSigma_0)$ as the base measure with $\vmu_0 \in \mathbb{R}^{D+1+K}$ and diagonal $\Sigma_0$. Each $\theta_i \in \mathbb{R}^{D+1+K}$ is a concatenated vector of $[\theta_i^{(0)}\in \mathbb{R}^D,\tilde{\theta}_i^{(0)}\in \mathbb{R},\theta_i^{(1)}\in \mathbb{R}^K]$ formed from the feature extractor weight-bias pairs with the corresponding weights of the softmax regression. In what follows, we will use ``batch'' to refer to a partition of the data. 

Next, for each $j=1,\ldots,J$ select a subset of the global atoms for batch $j$ via the Bernoulli process:
\begin{equation}
\label{eq:j_bern}
\Tp_j  := \sum_i \rb_{ji} \delta_{\rvtheta_i}\text{, where } 
\rb_{ji}|\rq_{i} \sim \text{Bern}(\rq_{i})\,\forall i.
\end{equation}
$\Tp_j$ is supported by atoms $\{\rvtheta_i : \rb_{ji} = 1, i=1,2,\ldots \}$, which represent the identities of the atoms (neurons) used by batch $j$. Finally, assume that observed local atoms are noisy measurements of the corresponding global atoms:
\begin{equation}
\label{eq:j_gauss}
\rvv_{jl}|\Tp_j \sim \mathcal{N}(\Tp_{jl}, \mSigma_j)\text{ for }l = 1,\ldots,L_j; \:\: L_j := \card(\Tp_j),
\end{equation}
with $\rvv_{jl}=[v_{jl}^{(0)},\tilde{v}_{jl}^{(0)},v_{jl}^{(1)}]$ being the weights, biases, and softmax regression weights corresponding to the $l$-th neuron of the $j$-th MLP trained with $L_j$ neurons on the data of batch $j$. 

Under this model, the key quantity to be inferred is the collection of random variables that \textbf{match} observed atoms (neurons) at any batch to the global atoms. We denote the collection of these random variables as $\{\mB^j\}_{j=1}^J$, where $\mB^j_{i,l}=1$ implies that $\Tp_{jl}=\theta_i$ (there is a one-to-one correspondence between $\{\rb_{ji}\}_{i=1}^\infty$ and $\mB^j$).

\paragraph{Maximum a posteriori estimation.} We now derive an algorithm for MAP estimation of global atoms for the model presented above. The objective function to be maximized is the posterior of $\{\theta_i\}_{i=1}^\infty$ and $\{\mB^j\}_{j=1}^J$:
\begin{align}
    \label{eq:map}
    \argmax\limits_{\{\rvtheta_i\},\{\mB^j\}}&P(\{\rvtheta_i\},\{\mB^j\}|\{\rvv_{jl}\}) \\\nonumber&\propto P(\{\rvv_{jl}\}|\{\rvtheta_i\},\{\mB^j\})P(\{\mB^j\})P(\{\rvtheta_i\}).
\end{align}
Note that the next proposition easily follows from Gaussian-Gaussian conjugacy:
\begin{proposition}\label{Prop:MAP}
Given $\{\mB^j\}$, the MAP estimate of $\{\rvtheta_i\}$ is given by
\begin{equation}
\label{eq:gaus_map_general}
\hat \rvtheta_i = \frac{\vmu_0/\sigma_0^2 + \sum_{j,l}\emB^j_{i,l}\rvv_{jl}/\sigma_{j}^2}{1/\sigma_0^2 + \sum_{j,l}\emB^j_{i,l}/\sigma_{j}^2}\text{ for }i=1,\ldots,L,
\end{equation}
where for simplicity we assume $\mSigma_0 = \mI\sigma^2_0$ and $\mSigma_j = \mI\sigma^2_j$. 
\end{proposition}
Using this fact we can cast optimization corresponding to \eqref{eq:map} with respect to only $\{\mB^j\}_{j=1}^J$. Taking the natural logarithm we obtain:
\begin{equation}
\label{eq:objective_multi_batch_simplified}
 \argmax\limits_{\{\mB^j\}}\frac{1}{2}\sum_i \frac{\left\|\frac{\vmu_0}{\sigma_0^2} + \sum_{j,l}\emB^j_{i,l}\frac{\rvv_{jl}}{\sigma_{j}}^2\right\|^2}{1/\sigma_0^2+\sum_{j,l}\emB^j_{i,l}/\sigma_{j}^2} + \log(P(\{\mB^j\}).
\end{equation}
We consider an iterative optimization approach: fixing all but one $\mB^j$ we find corresponding optimal assignment, then pick a new $j$ at random and proceed until convergence. In the following we will use notation $-j$ to denote ``all but $j$''. Let $L_{-j} = \max\{i: B^{-j}_{i,l} = 1\}$ denote number of active global weights outside of group $j$. We now rearrange the \emph{first} term of \eqref{eq:objective_multi_batch_simplified} by partitioning it into $i=1,\ldots,L_{-j}$ and $i=L_{-j}+1,\ldots,L_{-j}+L_j$. We are interested in solving for $\mB^j$, hence we can modify the objective function by subtracting terms independent of $\mB^j$ and noting that $\sum_l\emB^j_{i,l}\in\{0,1\}$, i.e. it is 1 if some neuron from batch $j$ is matched to global neuron $i$ and 0 otherwise:
\begin{align}
\nonumber
&\frac{1}{2}\sum_i \frac{\|\vmu_0/\sigma_0^2 + \sum_{j,l}\emB^j_{i,l}\rvv_{jl}/\sigma_{j}^2\|^2}{1/\sigma_0^2+\sum_{j,l}\emB^j_{i,l}/\sigma_{j}^2}=\\\nonumber
&\sum_{i=1}^{L_{-j}+L_j}\sum_{l=1}^{L_j} \emB^j_{i,l}\left(\frac{\|\vmu_0/\sigma_0^2 + \rvv_{jl}/\sigma_{j}^2 + \sum_{-j,l}\emB^j_{i,l}\rvv_{jl}/\sigma_{j}^2\|^2}{1/\sigma_0^2 + 1/\sigma_{j}^2 + \sum_{-j,l}\emB^j_{i,l}/\sigma_{j}^2} \right.\\\label{eq:objective_j_hungarian_param}
&\qquad \qquad \qquad\left.- \frac{\|\vmu_0/\sigma_0^2 + \sum_{-j,l}\emB^j_{i,l}\rvv_{jl}/\sigma_{j}^2\|^2}{1/\sigma_0^2+\sum_{-j,l}\emB^j_{i,l}/\sigma_{j}^2}\right).
\end{align}
Now we consider the \emph{second} term of \eqref{eq:objective_multi_batch_simplified}:
\begin{equation*}
\log P(\{\mB^j\}) = \log P(\mB^j|\mB^{-j}) + \log P(\mB^{-j}).
\end{equation*}
First, because we are optimizing for $\mB^j$, we can ignore $\log P(\mB^{-j})$. Second, due to exchangeability of batches (i.e. customers of the IBP), we can always consider $\mB^j$ to be the last batch (i.e. last customer of the IBP). Let $m^{-j}_i = \sum_{-j,l}\emB^j_{i,l}$ denote number of times batch weights were assigned to global weight $i$ outside of group $j$. We then obtain:
\begin{align}
\label{eq:objective_j_hungarian_nonparam}
\log P&(\{\mB^j\}) = \sum_{i=1}^{L_{-j}}\sum_{l=1}^{L_j} \emB^j_{i,l}\log\frac{m^{-j}_i}{J - m^{-j}_i} \\\nonumber
&+ \sum_{i=L_{-j} + 1}^{L_{-j} + L_j}\sum_{l=1}^{L_j} \emB^j_{i,l}\left(\log\frac{\gamma_0}{J} - \log(i-L_{-j})\right).
\end{align}
Combining \eqref{eq:objective_j_hungarian_param} and \eqref{eq:objective_j_hungarian_nonparam} we obtain the assignment cost objective, which we solve with the Hungarian algorithm.
\begin{proposition}\label{Prop:CostMatrix}
The (negative) assignment cost specification for finding $\mB^j$ is $- \emC^j_{i,l} = $
\begin{align}
\label{eq:cost_J}
& \begin{cases} \frac{\left\|\frac{\vmu_0}{\sigma_0^2} + \frac{\rvv_{jl}}{\sigma_{j}^2} + \sum\limits_{-j,l}\emB^j_{i,l}\frac{\rvv_{jl}}{\sigma_{j}^2}\right\|^2}{\frac{1}{\sigma_0^2} + \frac{1}{\sigma_{j}^2} + \sum_{-j,l}\emB^j_{i,l}/\sigma_{j}^2} - \frac{\left\|\frac{\vmu_0}{\sigma_0^2} + \sum\limits_{-j,l}\emB^j_{i,l}\frac{\rvv_{jl}}{\sigma_{j}^2}\right\|^2}{\frac{1}{\sigma_0^2}+\sum_{-j,l}\emB^j_{i,l}/\sigma_{j}^2} + 2\log\frac{m^{-j}_i}{J - m^{-j}_i},\\ &\hspace{-60pt} i \leq L_{-j} \\
\frac{\left\|\frac{\vmu_0}{\sigma_0^2} + \frac{\rvv_{jl}}{\sigma_{j}^2}\right\|^2}{\frac{1}{\sigma_0^2} + \frac{1}{\sigma_{j}^2}} -\frac{\left\|\frac{\vmu_0}{\sigma_0^2}\right\|^2}{\frac{1}{\sigma_0^2}}\hspace{-3pt} -\hspace{-3pt} 2\log\frac{i-L_{-j}}{\gamma_0/J} , & \hspace{-100pt} L_{-j} < i \leq L_{-j} + L_j.
\end{cases}
\end{align}
We then apply the Hungarian algorithm to find the \emph{minimizer} of $\sum_i\sum_l\emB^j_{i,l}\emC^j_{i,l}$ and obtain the neuron matching assignments. Proof is described in Supplement \ref{supp:SinglLayer}.
\end{proposition}
We summarize the overall single layer inference procedure in Figure \ref{Fig:DiagramOne} below.

\begin{figure}[t]
    \centering
    \noindent
    \begin{minipage}[t]{0.5\textwidth}
    \vspace{0pt}
    \begin{figure}[H]
    \centering
    \includegraphics[width=2.35in]{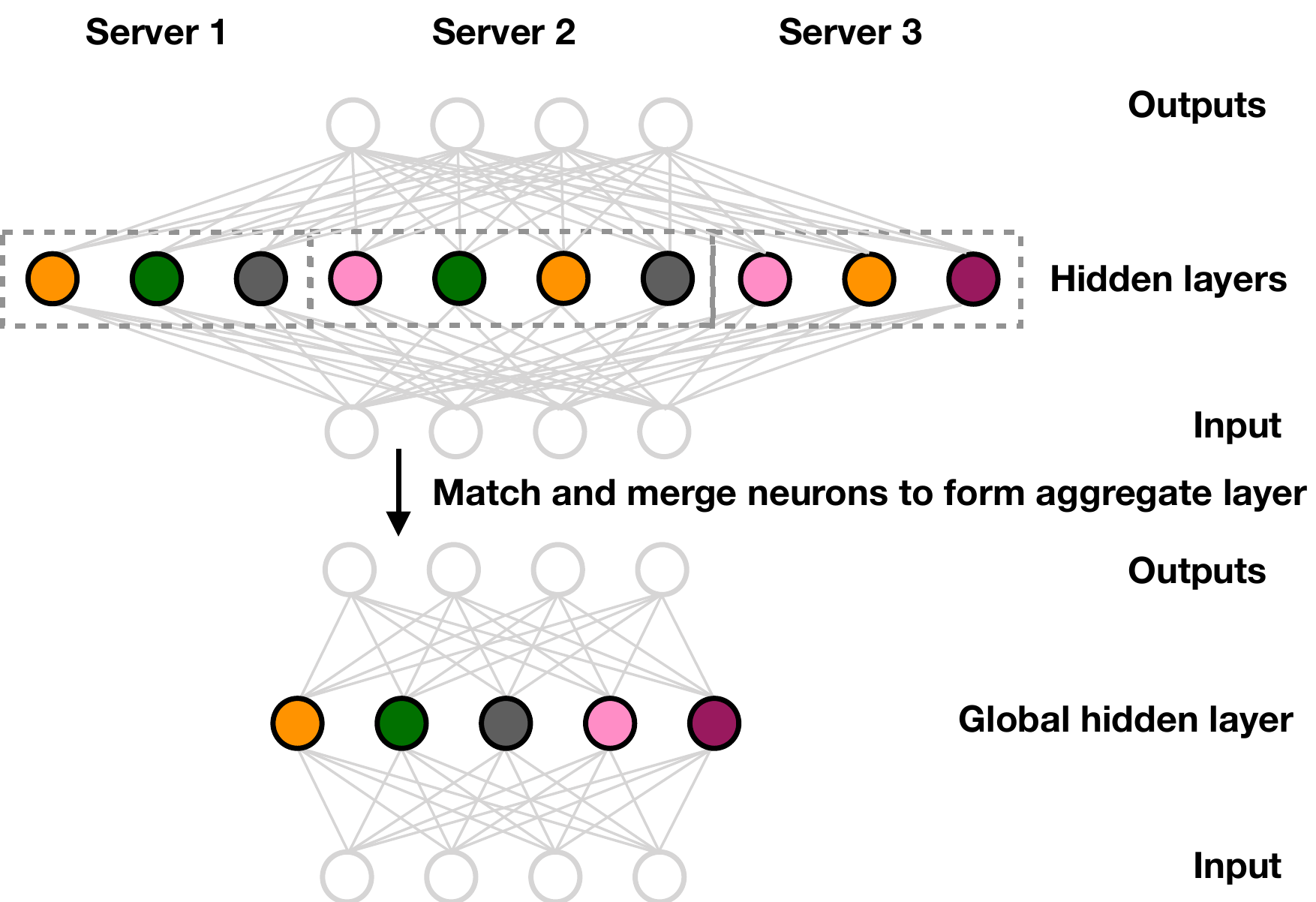}
    \end{figure}
    \end{minipage}
    \begin{minipage}[t]{0.45\textwidth}
    \vspace{0pt}
    \centering
    \begin{algorithm}[H]
\caption{Single Layer Neural Matching}
\begin{algorithmic}[1]
\STATE Collect weights and biases from the $J$ batches and form $\rvv_{jl}$. 
\STATE Form assignment cost matrix per \eqref{eq:cost_J}.
\STATE Compute matching assignments $B^j$ using the Hungarian algorithm.
\STATE Enumerate all resulting unique global neurons and use \eqref{eq:gaus_map_general} to infer the associated global weight vectors from all instances of the global neurons across the $J$ batches.
\STATE Concatenate the global neurons and the inferred weights and biases to form the new global hidden layer. 
\end{algorithmic}
\label{alg:OneLayer}
\end{algorithm}
    \end{minipage}
      \caption{\small{Single layer Probabilistic Federated Neural Matching algorithm showing matching of three MLPs. Nodes in the graphs indicate neurons, neurons of the same color have been matched. Our approach consists of using the corresponding neurons in the output layer to convert the neurons in each of the $J$ batches to weight vectors referencing the output layer. These weight vectors are then used to form a cost matrix, which the Hungarian algorithm then uses to do the matching. The matched neurons are then aggregated via Proposition \ref{Prop:MAP} to form the global model.}}
    \label{Fig:DiagramOne}
\end{figure}

\subsection{Multilayer Neural Matching}
\label{subsec:multilayer}
The model we have presented thus far can handle any arbitrary width single layer neural network, which is known to be theoretically sufficient for approximating any function of interest \citep{hornik1989multilayer}. However, deep neural networks with moderate layer widths are known to be beneficial both practically \citep{lecun2015deep} and theoretically \citep{poggio2017and}. We extend our neural matching approach to these deep architectures by defining a generative model of deep neural network weights from outputs back to inputs (top-down). Let $C$ denote the number of hidden layers and $L^c$ the number of neurons on the $c$-th layer. Then $L^{C+1}=K$ is the number of labels and $L^{0}=D$ is the input dimension. In the top down approach, we consider the global atoms to be vectors of outgoing weights from a neuron instead of weights forming a neuron as it was in the single hidden layer model. This change is needed to avoid base measures with unbounded dimensions.

Starting with the top hidden layer $c=C$, we generate each layer following a model similar to that used in the single layer case. For each layer we generate a collection of global atoms and select a subset of them for each batch using Beta-Bernoulli process construction. $L^{c+1}$ is the number of neurons on the layer $c+1$, which controls the dimension of the atoms in layer $c$.
\begin{definition}[Multilayer generative process]
Starting with layer $c = C$, generate (as in the single layer process)
\begin{align}
\label{eq:multi_bp}
& Q^c|\gamma_0^c,H^c,L^{c+1} \sim \mathrm{BP}(1, \gamma_0^c H^c),\\\nonumber
&\mathrm{\: then }\: Q^c = \sum_i q_i^c\delta_{\rvtheta_i^c},\ \rvtheta_i^c \sim \mathcal{N}(\vmu_0^c,\mSigma_0^c),\ \vmu_0^c \in \mathbb{R}^{L^{c+1}}\\\nonumber
& \Tp_j^c  := \sum_i \rb_{ji}^c \delta_{\rvtheta_i^c},\mathrm{ \: where } \:\:
\rb_{ji}^c|\rq_{i}^c \sim \mathrm{Bern}(\rq_{i}^c).
\end{align}
This $\mathcal{T}_j^c$ is the set of global atoms (neurons) used by batch $j$ in layer $c$, it contains atoms $\{\rvtheta_i^c: \rb_{ji}^c = 1, i=1,2,\ldots \}$. Finally, generate the observed local atoms:
\begin{equation}
\label{eq:j_c_gaussMulti}
\rvv_{jl}^c|\Tp_j^c, \sim \mathcal{N}(\Tp_{jl}^c, \mSigma_j^c)\text{ for }l = 1,\ldots,L_j^c,
\end{equation}
 where we have set $L_j^c := \card(\Tp_j^c)$.
Next, compute the generated number of global neurons $L^c = \mathrm{card}\{\cup_{j =1}^J \Tp_j^c\}$ and repeat this generative process for the next layer $c-1$. Repeat until all layers are generated ($c = C,\dots, 1$). 
\end{definition}
An important difference from the single layer model is that we should now set to 0 some of the dimensions of $\rvv_{jl}^c \in \mathbb{R}^{L^{c+1}}$ since they correspond to weights outgoing to neurons of the layer $c+1$ not present on the batch $j$, i.e. $\rvv_{jli}^c := 0$ if $\rb_{ji}^{c+1}=0$ for $i=1,\ldots,L^{c+1}$. The resulting model can be understood as follows. There is a global fully connected neural network with $L^c$ neurons on layer $c$ and there are $J$ partially connected neural networks with $L^c_j$ active neurons on layer $c$, while weights corresponding to the remaining $L^c-L^c_j$ neurons are zeroes and have no effect locally.

\begin{remark}
Our model can conceptually handle permuted ordering of the input dimensions across batches, however in most practical cases the ordering of input dimensions is consistent across batches. Thus, we assume that the weights connecting the first hidden layer to the inputs exhibit permutation invariance only on the side of the first hidden layer. Similarly to how all weights were concatenated in the single hidden layer model, we consider $\vmu_0^c \in \mathbb{R}^{D+L^{c+1}}$ for $c=1$. We also note that the bias term can be added to the model, we omitted it to simplify notation.
\end{remark} 

\paragraph{Inference}
Following the top-down generative model, we adopt a greedy inference procedure that first infers the matching of the top layer and then proceeds down the layers of the network. This is possible because the generative process for each layer depends only on the identity and number of the global neurons in the layer above it, hence once we infer the $c+1$th layer of the global model we can apply the single layer inference algorithm (Algorithm \ref{alg:OneLayer}) to the $c$th layer. This greedy setup is illustrated in Supplement Figure \ref{Fig:Diagram}. The per-layer inference follows directly from the single layer case, yielding the following propositions.
\begin{proposition}
The (negative) assignment cost specification for finding $\mB^{j,c}$ is $-\emC^{j,c}_{i,l}=$
\begin{align*}
\label{eq:cost_JMulti}
\begin{cases} \begin{aligned}[b]& \frac{\left\|\frac{\vmu_0^c}{(\sigma_0^c)^2} + \frac{\rvv_{jl}^c}{(\sigma_{j}^c)^2} + \sum\limits_{-j,l}\emB^{j,c}_{i,l}\frac{\rvv_{jl}^c}{(\sigma_{j}^c)^2}\right\|^2}{\frac{1}{(\sigma_0^c)^2} + \frac{1}{(\sigma_{j}^c)^2} + \sum_{-j,l}\emB^{j,c}_{i,l}/(\sigma_{j}^c)^2} + 2\log\frac{m^{-j,c}_i}{J - m^{-j,c}_i}
\\&\:\:- \frac{\left\|{\vmu_0^c}/{(\sigma_0^c)^2} + \sum_{-j,l}\emB^{j,c}_{i,l}{\rvv_{jl}^c}/{(\sigma_{j}^c)^2}\right\|^2}{{1}/{(\sigma_0^c)^2}+\sum_{-j,l}\emB^{j,c}_{i,l}/(\sigma_{j}^c)^2}, \end{aligned}, & \hspace{-50pt}i \leq L_{-j}^c \\
\begin{aligned}[b]&\frac{\left\|\frac{\vmu_0^c}{(\sigma_0^c)^2} + \frac{\rvv_{jl}^c}{(\sigma_{j}^c)^2}\right\|^2}{\frac{1}{(\sigma_0^c)^2} + \frac{1}{(\sigma_{j}^c)^2}} -\frac{\left\|{\vmu_0^c}/{(\sigma_0^c)^2}\right\|^2}{1/(\sigma_0^c)^2} - 2\log\frac{i-L_{-j}^c}{\gamma_0/J} ,\\\\\end{aligned} & \hspace{-100pt} L_{-j}^c < i \leq L_{-j}^c + L_j^c,
\end{cases}
\end{align*}

where for simplicity we assume $\mSigma_0^c = \mI(\sigma^c_0)^2$ and $\mSigma_j^c = \mI(\sigma^c_j)^2$. 
We then apply the Hungarian algorithm to find the \emph{minimizer} of $\sum_i\sum_l\emB^{j,c}_{i,l}\emC^{j,c}_{i,l}$ and obtain the neuron matching assignments.
\end{proposition}
\begin{proposition}
Given the assignment $\{\mB^{j,c}\}$, the MAP estimate of $\{\rvtheta_i^c\}$ is given by
\begin{equation}
\label{eq:gaus_map_generalMulti}
\hat \rvtheta_i^c = \frac{\vmu_0^c/(\sigma_0^c)^2 + \sum_{j,l}\emB^{j,c}_{i,l}\rvv_{jl}^c/(\sigma_{j}^c)^2}{1/(\sigma_0^c)^2 + \sum_{j,l}\emB^{j,c}_{i,l}/(\sigma_{j}^c)^2}\text{ for }i=1,\ldots,L.
\end{equation}
\end{proposition}
We combine these propositions and summarize the overall multilayer inference procedure in Supplement Algorithm \ref{alg:MultiLayer}.

\subsection{Neural Matching with Additional Communications}
\label{subsec:communication}
In the traditional federated learning scenario, where local and global models are learned together, common approach (see e.g., \citet{mcmahan2017communication}) is to learn via rounds of communication between local and global models. Typically, local model parameters are trained for few epochs, sent to server for updating the global model and then reinitialized with the global model parameters for the new round. One of the key factors in federated learning is the number of communications required to achieve accurate global model. In the preceding sections we proposed Probabilistic Federated Neural Matching (PFNM) to aggregate local models in a single communication round. Our approach can be naturally extended to benefit from additional communication rounds as follows.

Let $t$ denote a communication round. To initialize local models at round $t+1$ we set $\rvv_{jl}^{t+1} = \sum_i B_{i,l}^{j,t} \rvtheta_i^t$. Recall that $\sum_i B_{i,l}^{j,t} = 1 \ \forall l=1,\ldots,L_j,\ j=1,\ldots,J$, hence a local model is initialized with a \emph{subset} of the global model, keeping local model size $L_j$ constant across communication rounds (this also holds for the multilayer case). After local models are updated we proceed to apply matching to obtain new global model. Note that global model size can change across communication rounds, in particular we expect it to shrink as local models improve on each step.

\section{Experiments}
\label{sec:experiments}
To verify our methodology we simulate federated learning scenarios using two standard datasets: MNIST and CIFAR-10. We randomly partition each of these datasets into $J$ batches. Two partition strategies are of interest: (a) a homogeneous partition where each batch has approximately equal proportion of each of the $K$ classes; and (b) a heterogeneous partition for which batch sizes and class proportions are unbalanced. 
We simulate a heterogeneous partition by simulating $\rvp_k \sim \text{Dir}_J(0.5)$ and allocating a $\rvp_{k,j}$ proportion of the instances of class $k$ to batch $j$. Note that due to the small concentration parameter ($0.5$) of the Dirichlet distribution, some sampled batches may not have any examples of certain classes of data. For each of the four combinations of partition strategy and dataset we run $10$ trials to obtain mean performances with standard deviations.

In our empirical studies below, we will show that our framework can aggregate multiple local neural networks trained independently on different batches of data into an efficient, modest-size global neural network with as few as a single communication round. We also demonstrate enhanced performance when additional communication is allowed.

\begin{figure*}[t!]
\begin{subfigure}{.245\textwidth}
  \centering
  \captionsetup{justification=centering}
\includegraphics[width=\linewidth]{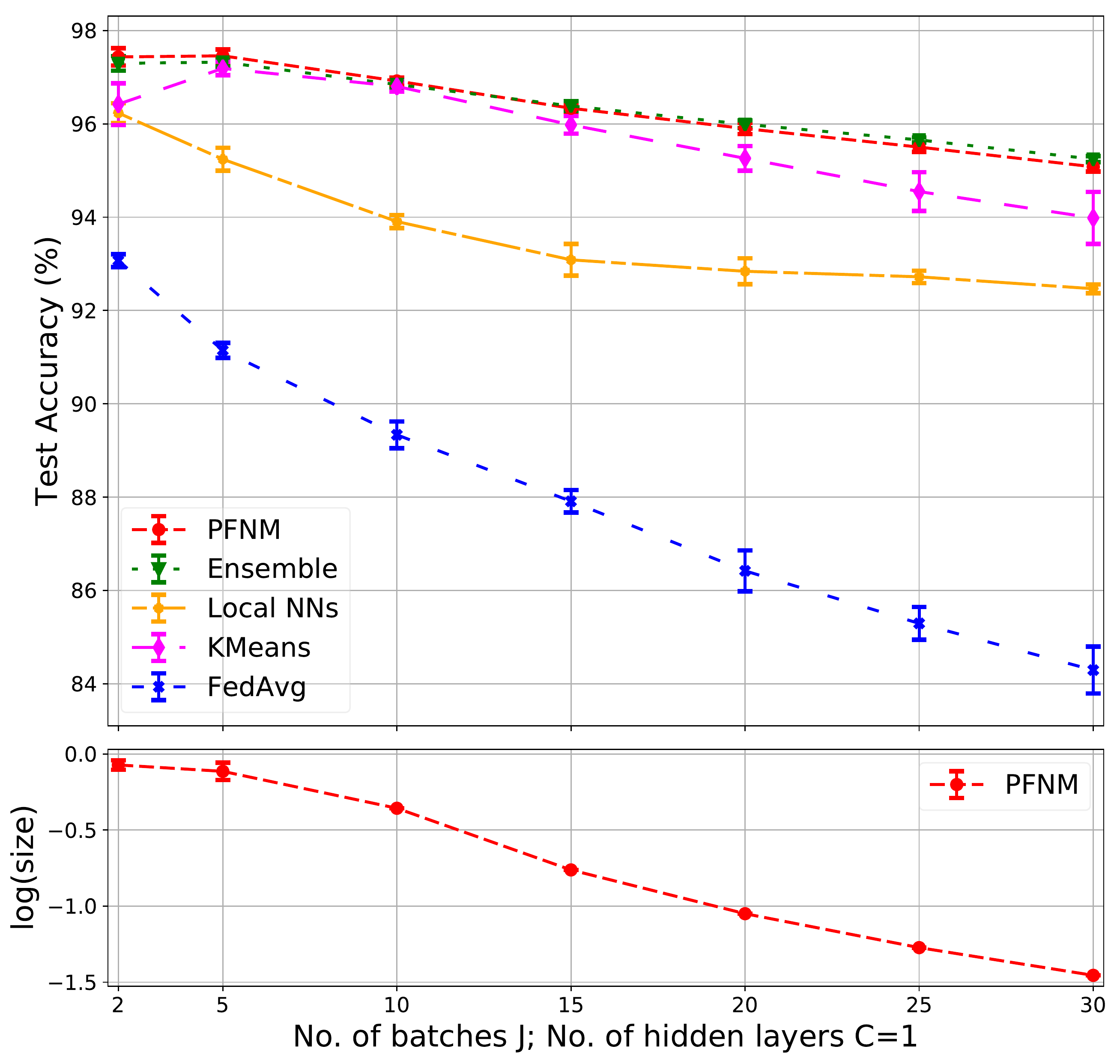}
\caption{MNIST homogeneous}
\label{fig:J100_mnist_homo_acc}
\end{subfigure}
\begin{subfigure}{.245\textwidth}
  \centering
  \captionsetup{justification=centering}
\includegraphics[width=\linewidth]{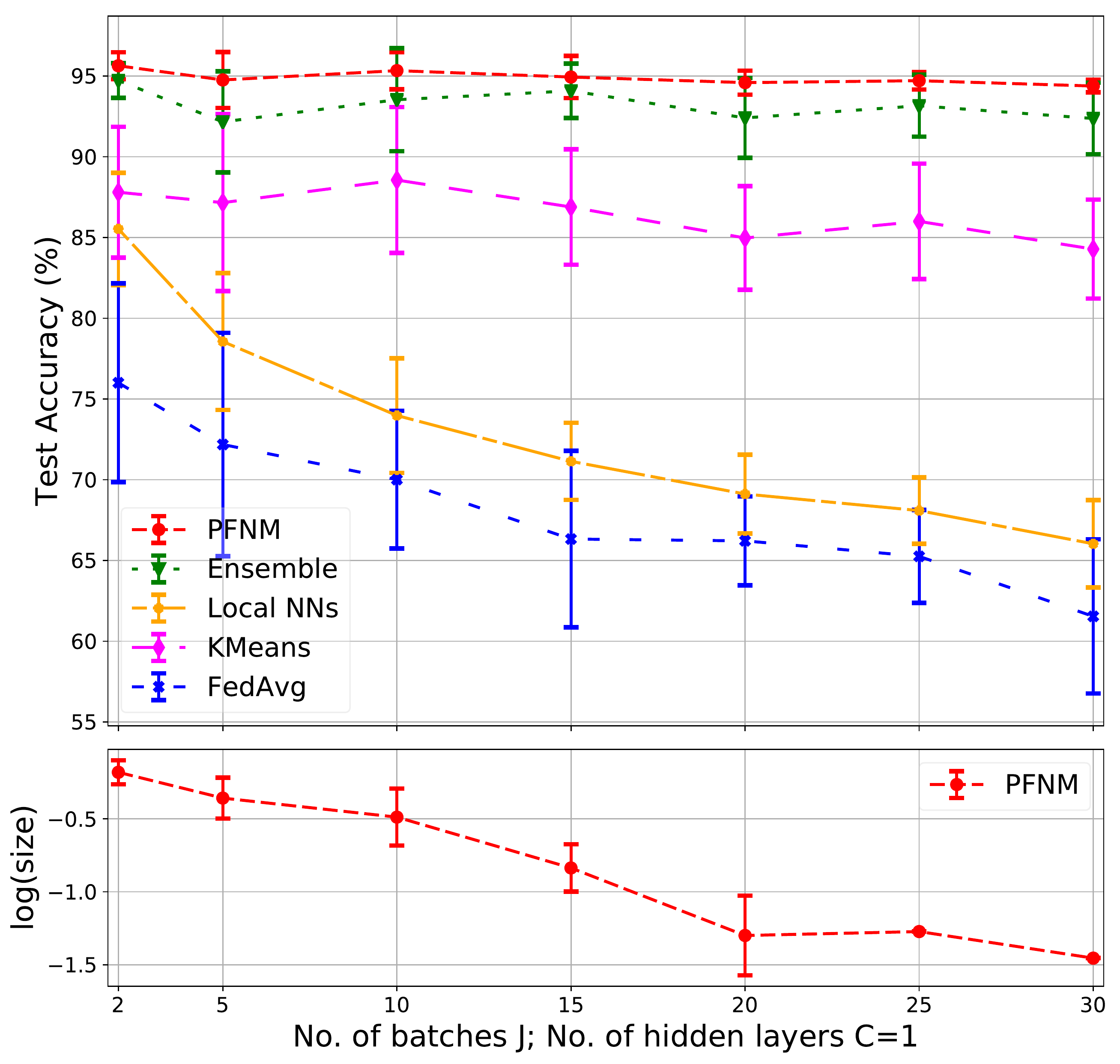}
\caption{MNIST heterogeneous}
\label{fig:J100_mnist_homo_shape}
\end{subfigure}
\begin{subfigure}{.245\textwidth}
  \centering
  \captionsetup{justification=centering}
\includegraphics[width=\linewidth]{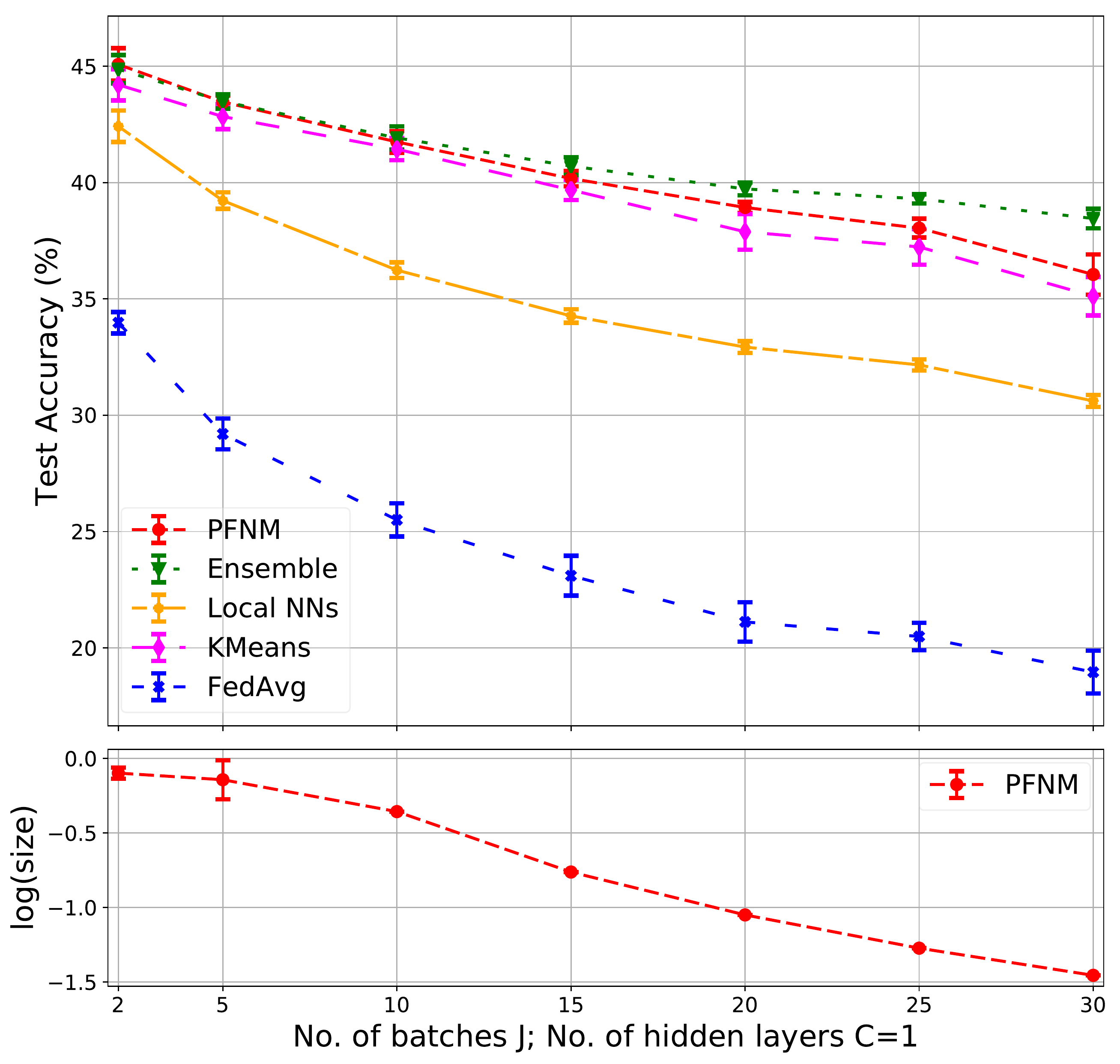}
\caption{CIFAR10 homogeneous}
\label{fig:J100_mnist_hetero_acc}
\end{subfigure}
\begin{subfigure}{.245\textwidth}
  \centering
  \captionsetup{justification=centering}
\includegraphics[width=\linewidth]{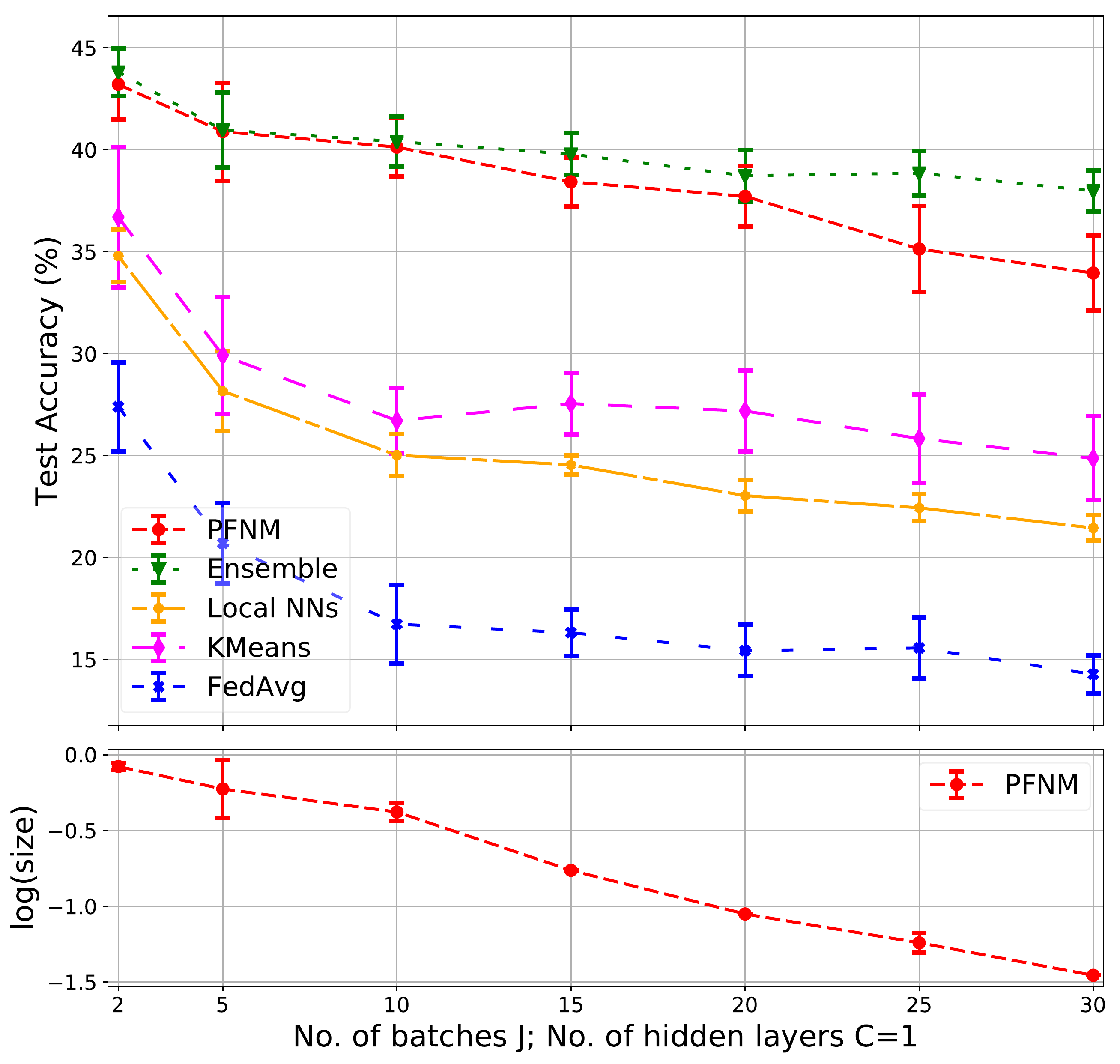}
\caption{CIFAR10 heterogeneous}
\label{fig:J100_mnist_hetero_shape}
\end{subfigure} \\
\begin{subfigure}{.245\textwidth}
  \centering
  \captionsetup{justification=centering}
\includegraphics[width=\linewidth]{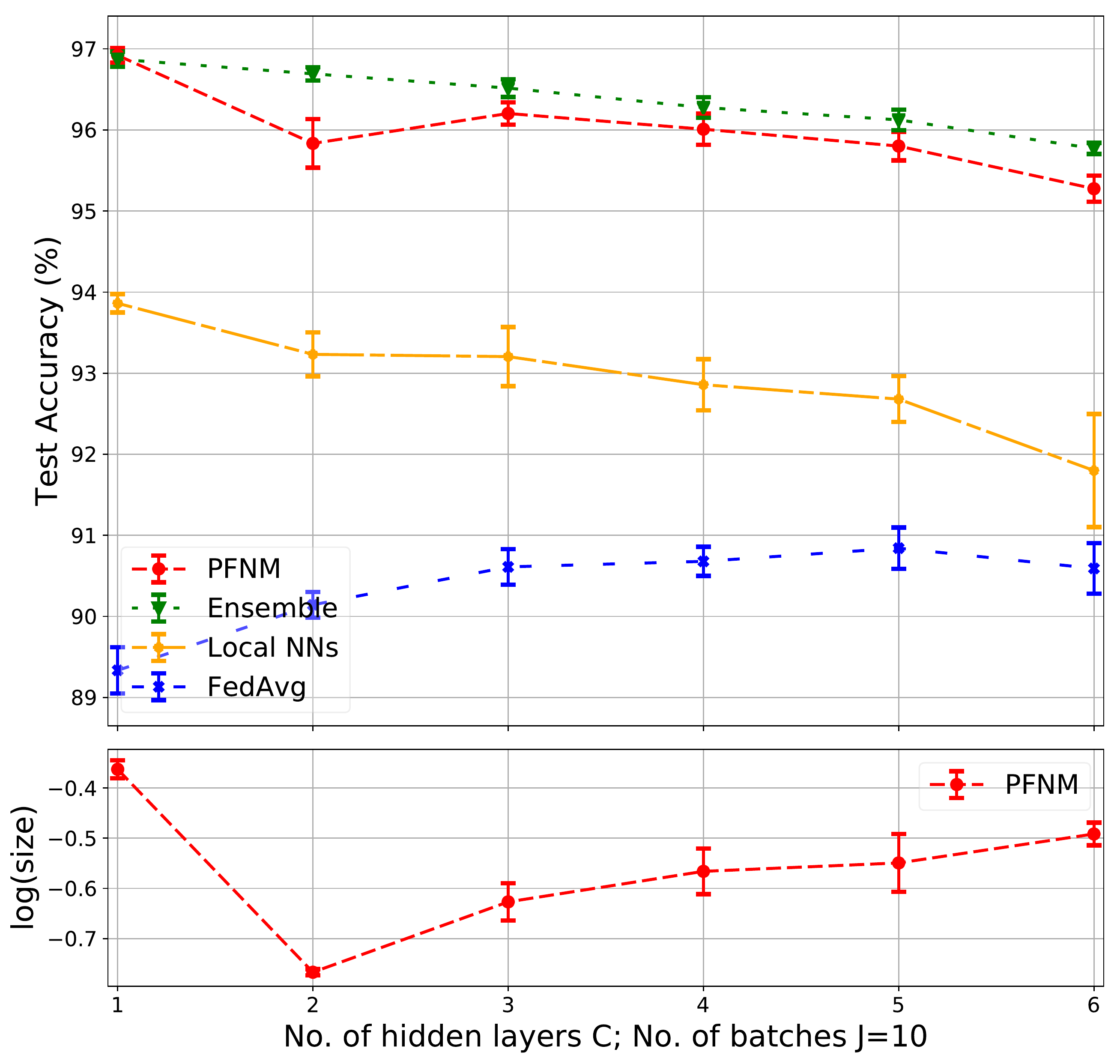}
\caption{MNIST homogeneous}
\label{fig:C_mnist_homo}
\end{subfigure}
\begin{subfigure}{.245\textwidth}
  \centering
  \captionsetup{justification=centering}
\includegraphics[width=\linewidth]{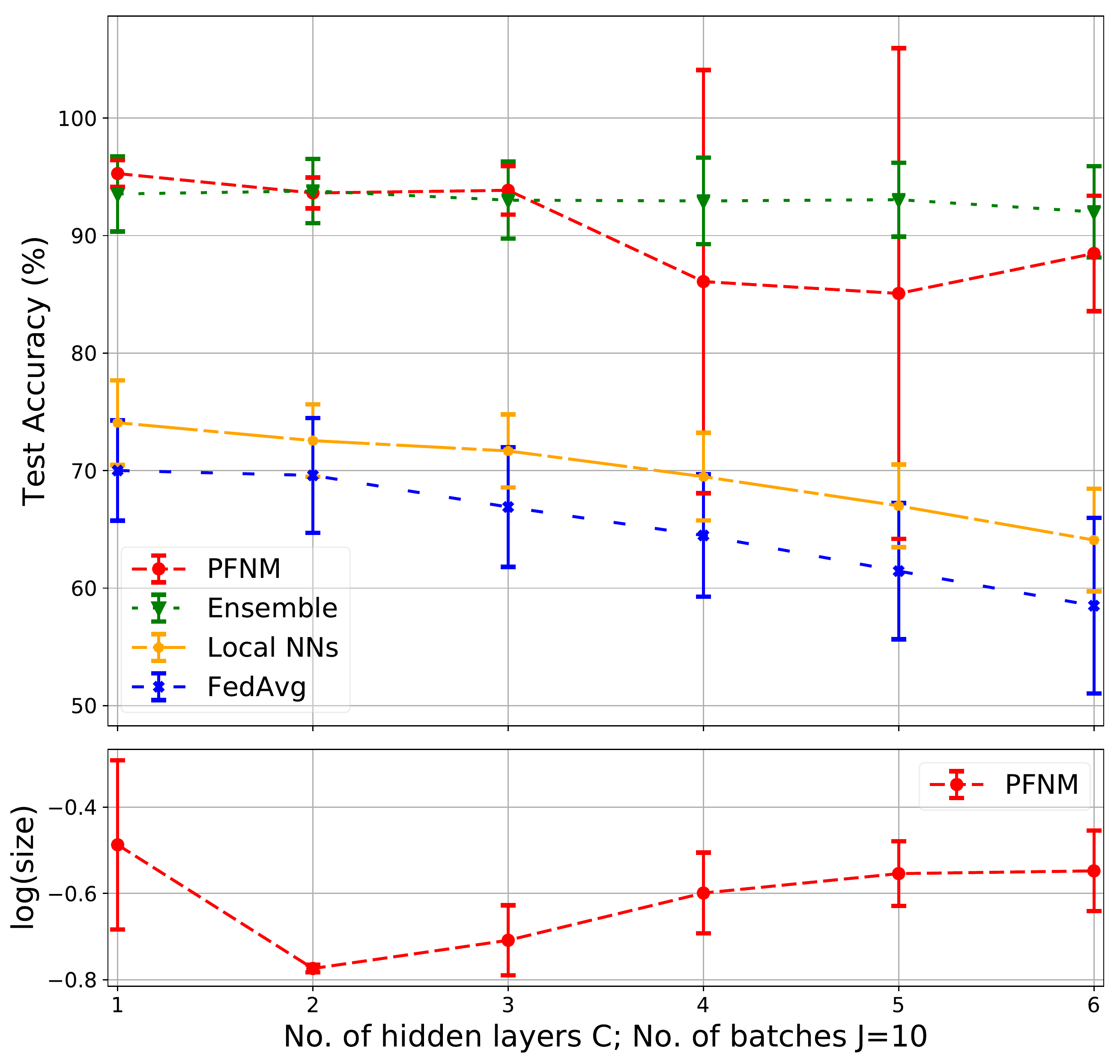}
\caption{MNIST heterogeneous}
\label{fig:C_mnist_hetero}
\end{subfigure}
\begin{subfigure}{.245\textwidth}
  \centering
  \captionsetup{justification=centering}
\includegraphics[width=\linewidth]{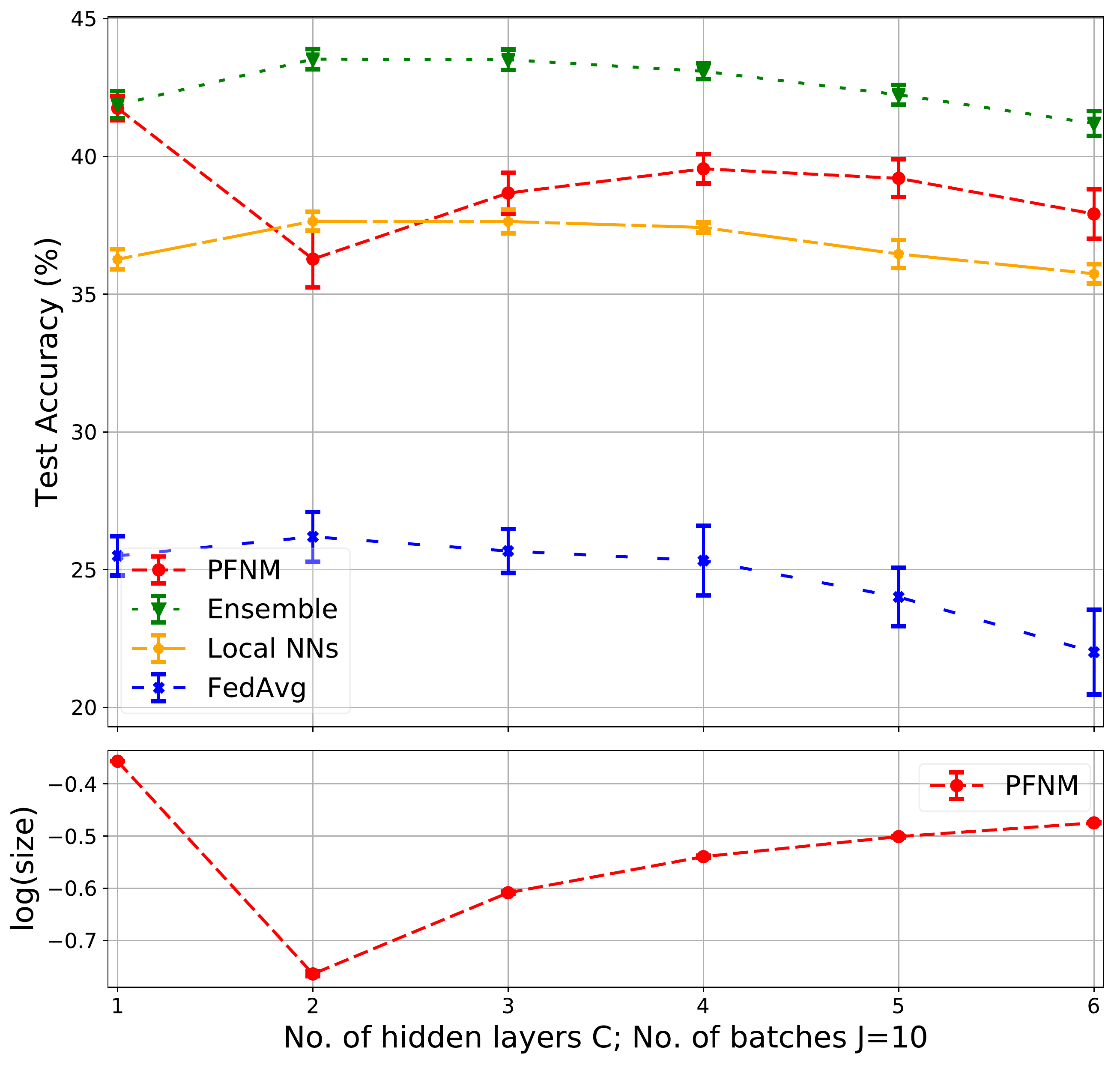}
\caption{CIFAR homogeneous}
\label{fig:C_cifar_homo}
\end{subfigure}
\begin{subfigure}{.245\textwidth}
  \centering
  \captionsetup{justification=centering}
\includegraphics[width=\linewidth]{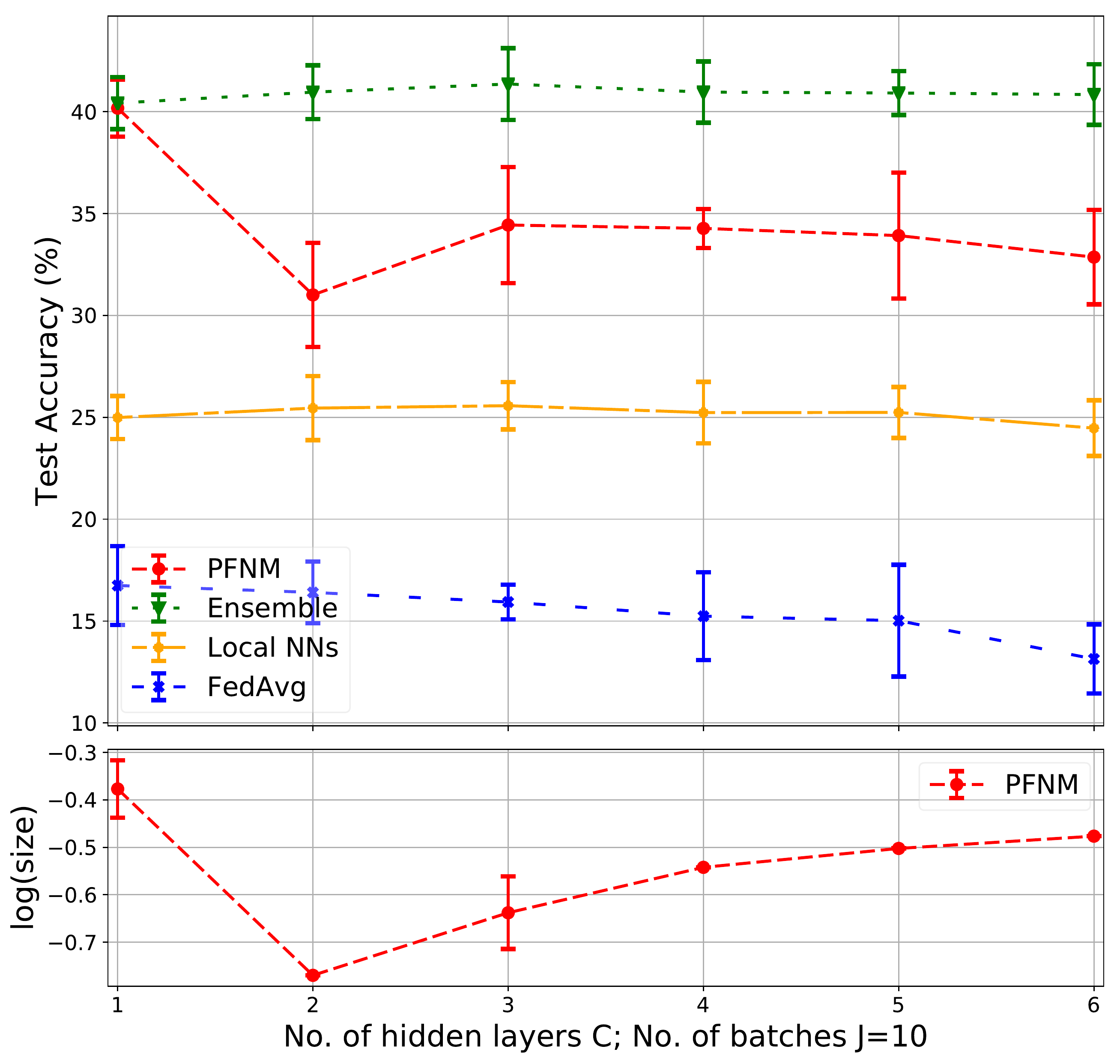}
\caption{CIFAR heterogeneous}
\label{fig:C_cifar_hetero}
\end{subfigure}
\caption{{\textbf{Single communication federated learning}. TOP: Test accuracy and normalized model size ($\log \frac{L}{\sum_j L_j}$) as a function of varying number of batches ($J$). BOTTOM: Test accuracy and normalized model size for multi-layer networks as a function of number of layers. PFNM consistently outperforms local models and federated averaging while performing comparably to ensembles at a fraction of the storage and computational costs.}}
\label{fig:J100}
\end{figure*}

\paragraph{Learning with single communication}
First we consider a scenario where a global neural network needs to be constructed with a single communication round. This imitates the real-world scenario where data is no longer available and we only have access to pre-trained local models (i.e. ``legacy'' models). To be useful, this global neural network needs to outperform the individual local models. Ensemble methods~\citep{dietterich2000ensemble, breiman2001random} are a classic approach for combining predictions of multiple learners. They often perform well in practice even when the ensemble members are of poor quality. Unfortunately, in the case of neural networks, ensembles have large storage and inference costs, stemming from having to store and forward propagate through all local networks.   

The performance of local NNs and the ensemble method define the lower and upper extremes of aggregating when limited to a single communication. We also compare to other strong baselines, including federated averaging of local neural networks trained with the same random initialization as proposed by \citet{mcmahan2017communication}. We note that a federated averaging variant without the shared initialization would likely be more realistic when trying to aggregate pre-trained models, but this variant performs significantly worse than all other baselines. We also consider k-Means clustering \citep{lloyd1982least} of vectors constructed by concatenating weights and biases of local neural networks. The key difference between k-Means and our approach is that clustering, unlike matching, allows several neurons from a single neural network to be assigned to the same global neuron, potentially averaging out their individual feature representations. Further, k-Means requires us to choose k, which we set to $K=\min(500, 50J)$. In contrast, PFNM nonparametrically learns the global model size and other hyperparameters, i.e. $\sigma, \sigma_0, \gamma_0$, are chosen based on the training data. We discuss parameter sensitivity in Supplement \ref{supp:param_sensitivity}.


Figure \ref{fig:J100} presents our results with single hidden layer neural networks for varying number of batches $J$. Note that a higher number of batches implies fewer data instances per batch, leading to poorer local model performances. The upper plots summarize test data accuracy, while the lower plots show the model size compression achieved by PFNM. Specifically we plot $\log \frac{L}{\sum_j L_j}$, which is the log ratio of the PFNM global model size $L$ to the total number of neurons across all local models (i.e. the size of an ensemble model). In this and subsequent experiments each local neural network has $L_j = 100$ hidden neurons. We see that PFNM produces strong results, occasionally even outperforming ensembles. In the heterogeneous setting we observe a noticeable degradation in the performance of the local NNs and of k-means, while PFNM retains its good performance. It is worth noting that the gap between PFNM and ensemble increases on CIFAR10 with $J$, while it is constant (and even in favor of PFNM) on MNIST. This is not surprising. Ensemble methods are known to perform particularly well at aggregating ``weak'' learners (recall higher $J$ implies smaller batches)~\citep{breiman2001random}, while PFNM assumes the neural networks being aggregated already perform reasonably well.

Next, we investigate aggregation of multi-layer neural networks, each using a hundred neurons per layer. The extension of k-means to this setting is unclear and k-means is excluded from further comparisons. In Figure \ref{fig:J100}, we show that PFNM again provides drastic and consistent improvements over local models and federated averaging. It performs marginally worse than ensembles, especially for deeper networks on CIFAR10. This aligns with our previous observation --- when there is insufficient data for training good local models, PFNM's performance marginally degrades with respect to ensembles, but still provides significant compression over ensembles.  

We briefly discuss complexity of our algorithms and experiment run-times in Supplement \ref{supp:complexity}.

\begin{figure*}[t!]
\begin{subfigure}{.245\textwidth}
  \centering
  \captionsetup{justification=centering}
\includegraphics[width=\linewidth]{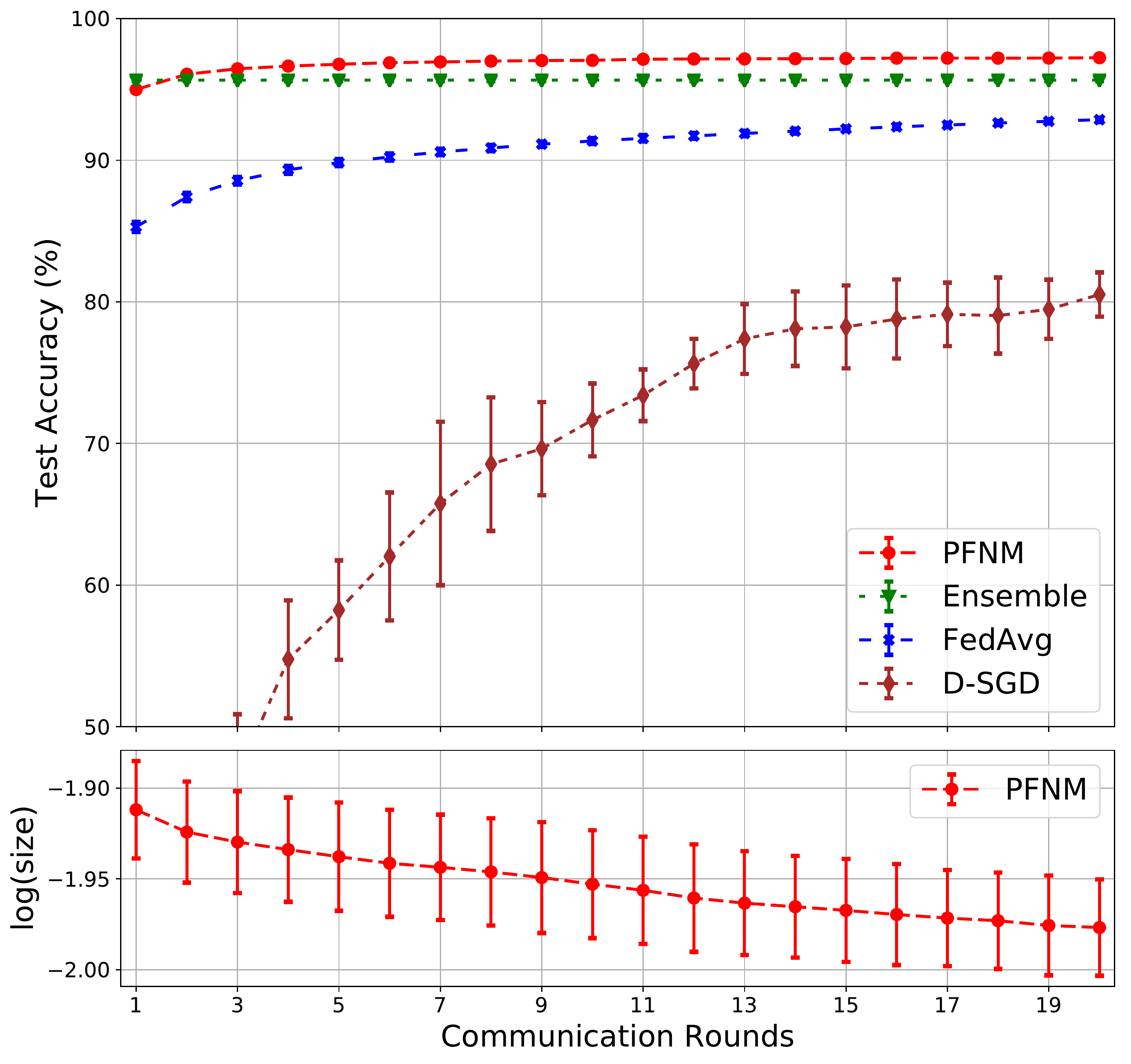}
\caption{MNIST homogeneous}
\label{fig:mnist_1_homo_comm}
\end{subfigure}
\begin{subfigure}{.245\textwidth}
  \centering
  \captionsetup{justification=centering}
\includegraphics[width=\linewidth]{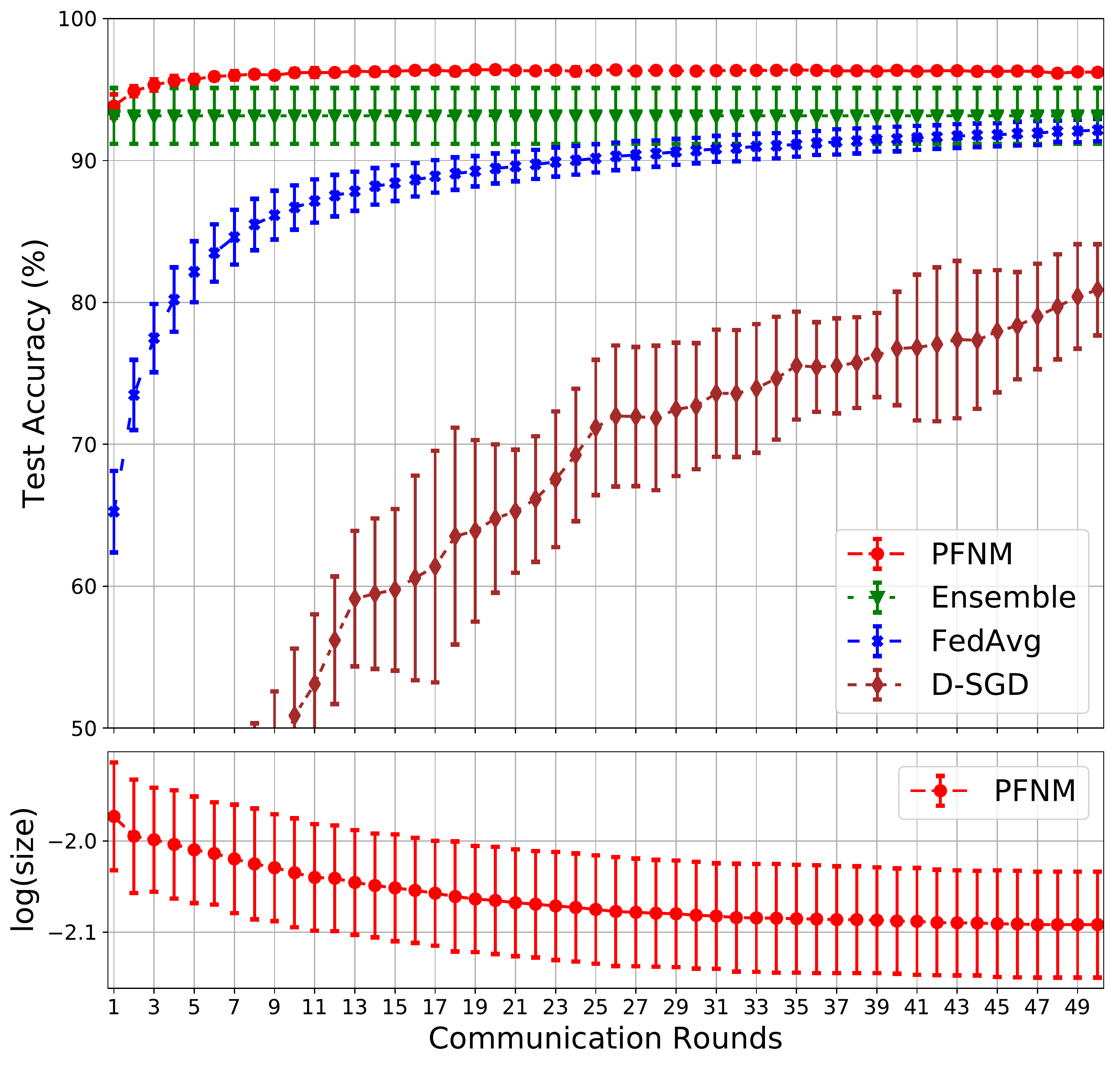}
\caption{MNIST heterogeneous}
\label{fig:mnist_1_hetero_comm}
\end{subfigure}
\begin{subfigure}{.245\textwidth}
  \centering
  \captionsetup{justification=centering}
\includegraphics[width=\linewidth]{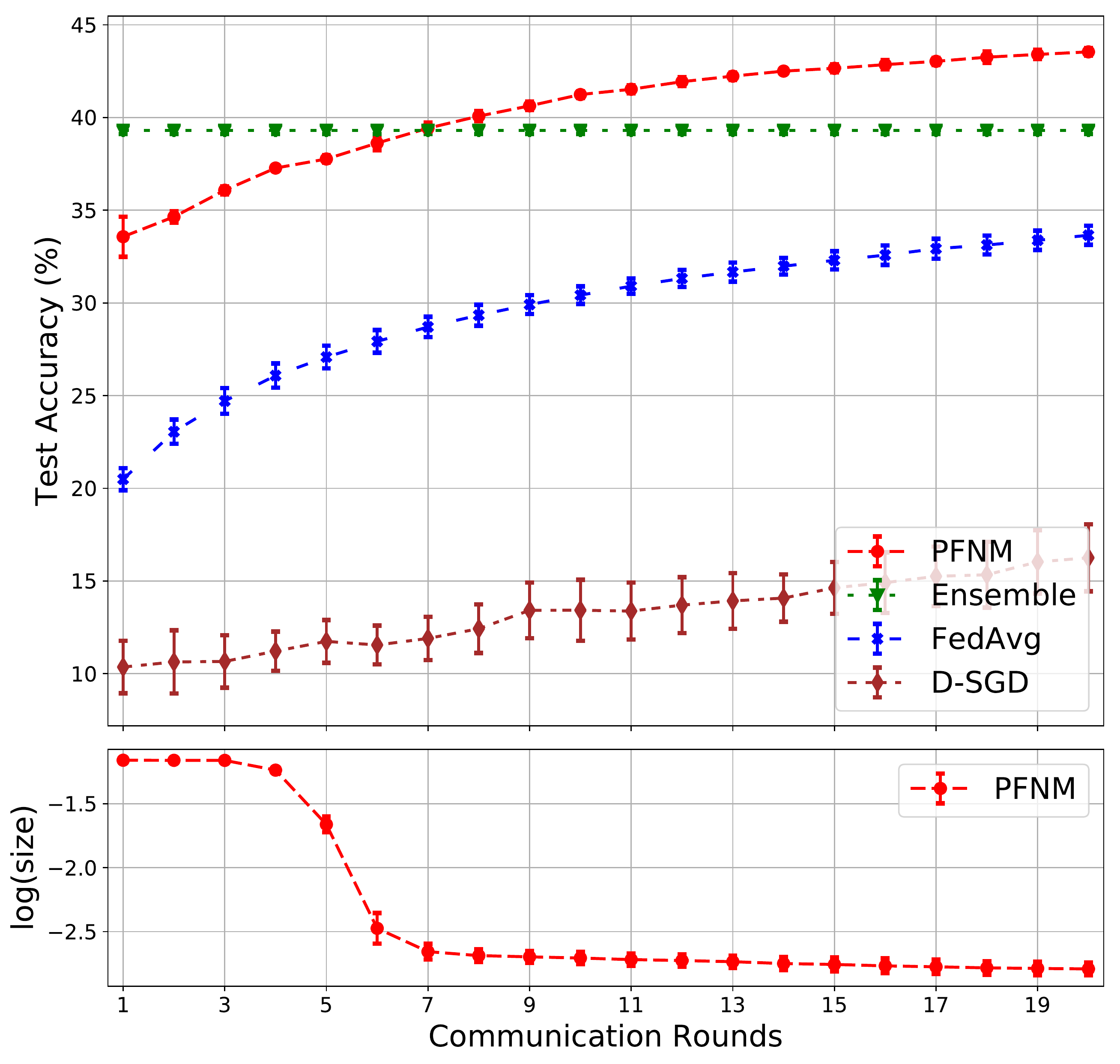}
\caption{CIFAR homogeneous}
\label{fig:cifar_1_homo_comm}
\end{subfigure}
\begin{subfigure}{.245\textwidth}
  \centering
  \captionsetup{justification=centering}
\includegraphics[width=\linewidth]{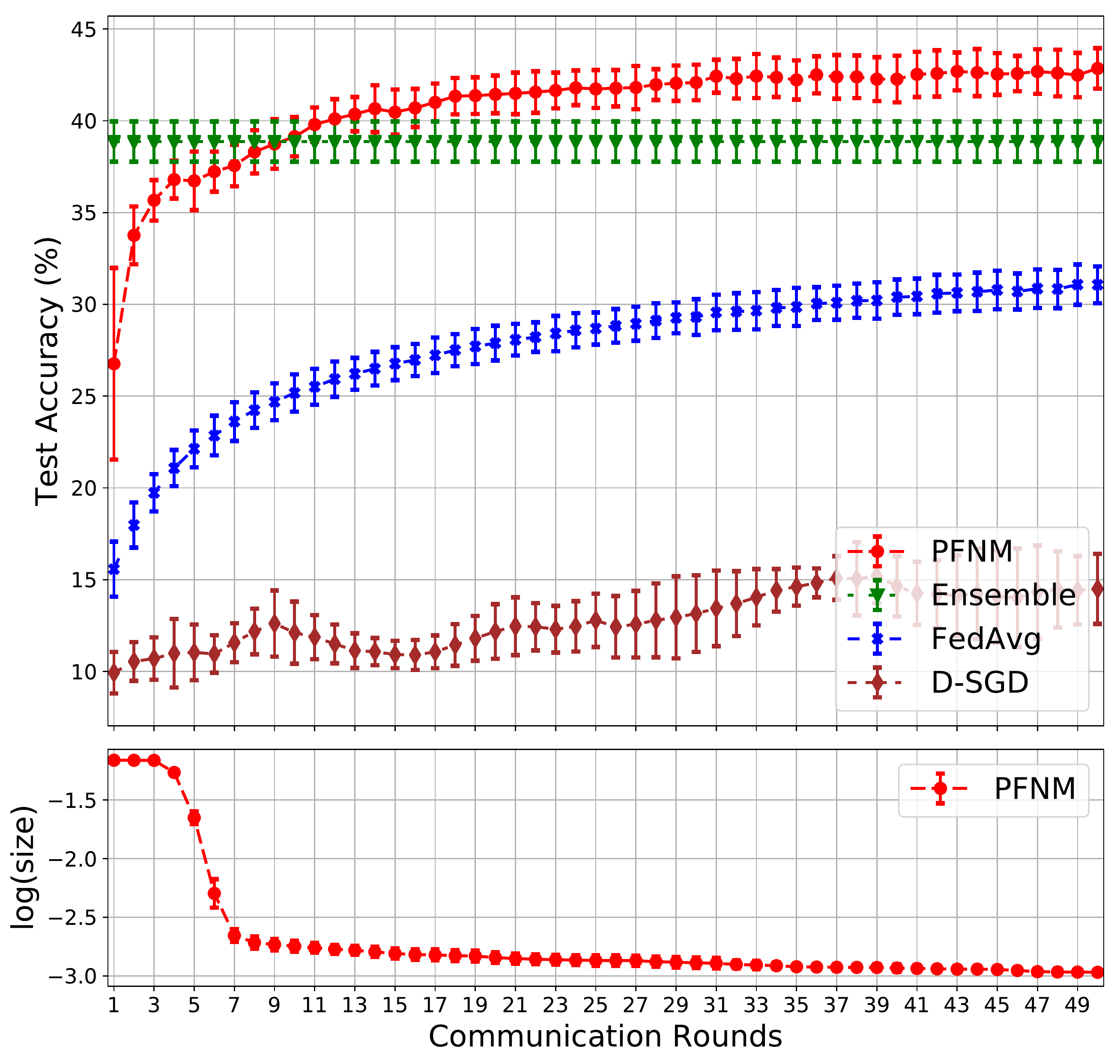}
\caption{CIFAR heterogeneous}
\label{fig:cifar_1_hetero_comm}
\end{subfigure} \\
\begin{subfigure}{.245\textwidth}
  \centering
  \captionsetup{justification=centering}
\includegraphics[width=\linewidth]{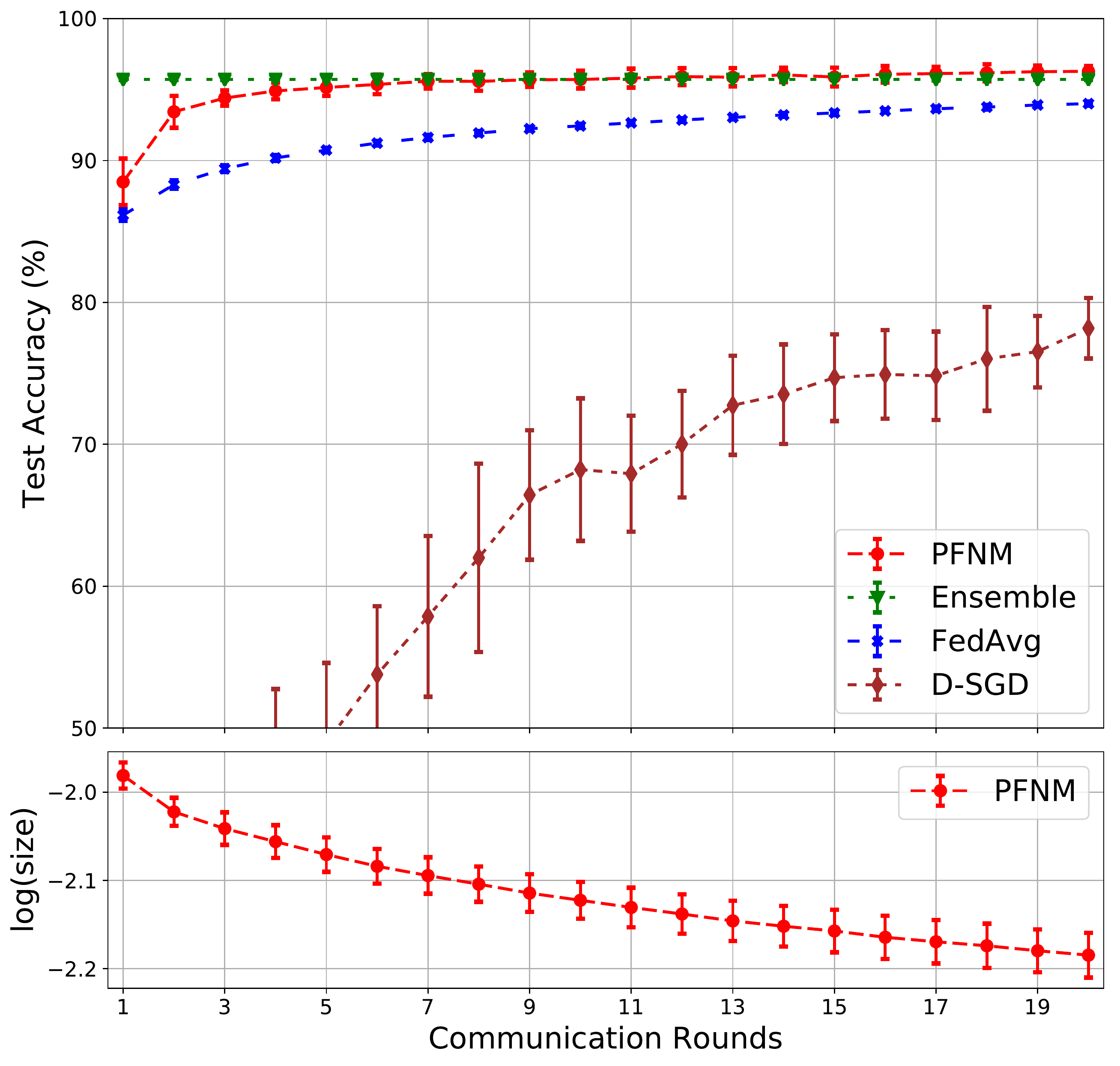}
\caption{MNIST homogeneous}
\label{fig:mnist_2_homo_comm}
\end{subfigure}
\begin{subfigure}{.245\textwidth}
  \centering
  \captionsetup{justification=centering}
\includegraphics[width=\linewidth]{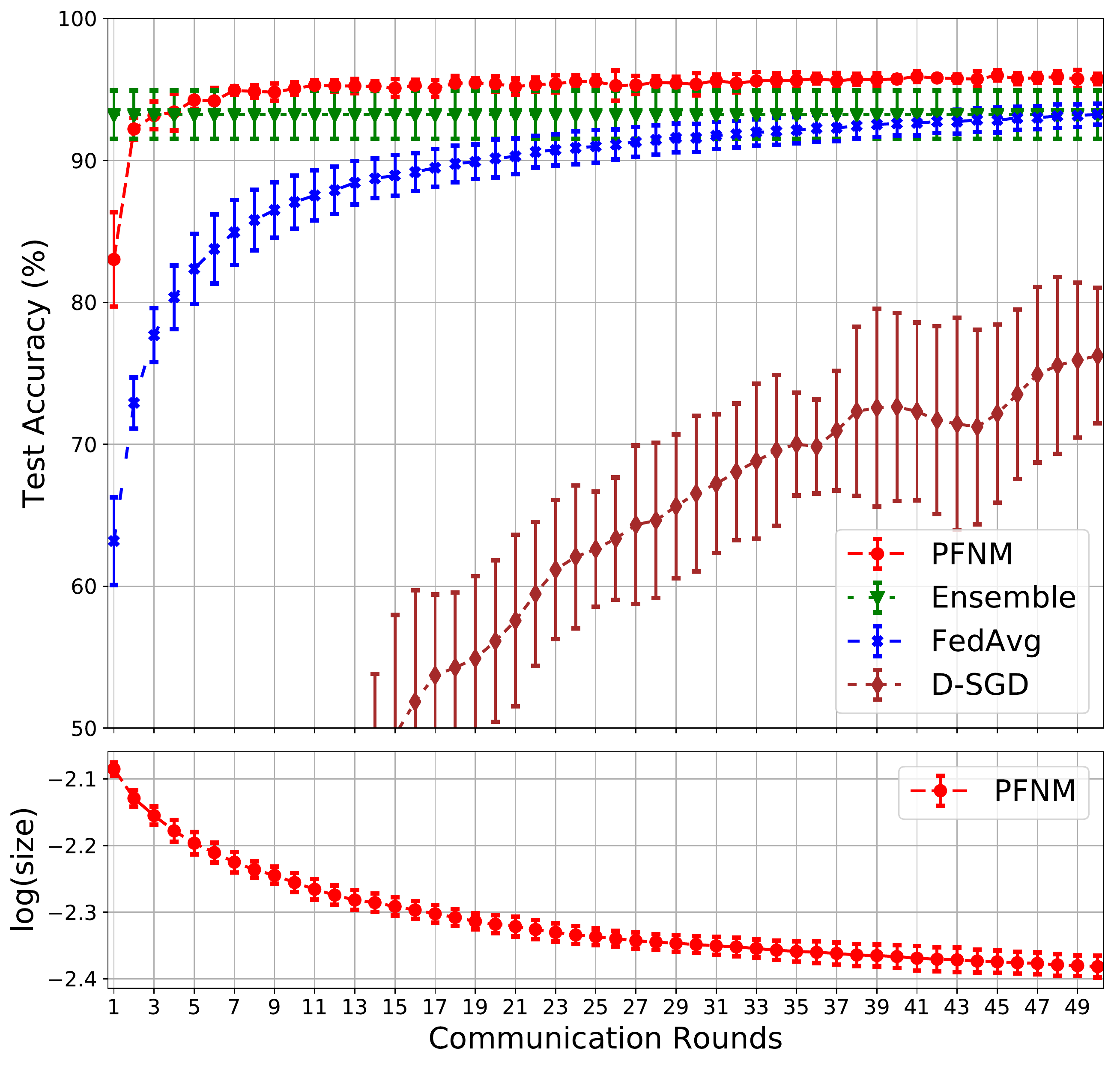}
\caption{MNIST heterogeneous}
\label{fig:mnist_2_hetero_comm}
\end{subfigure}
\begin{subfigure}{.245\textwidth}
  \centering
  \captionsetup{justification=centering}
\includegraphics[width=\linewidth]{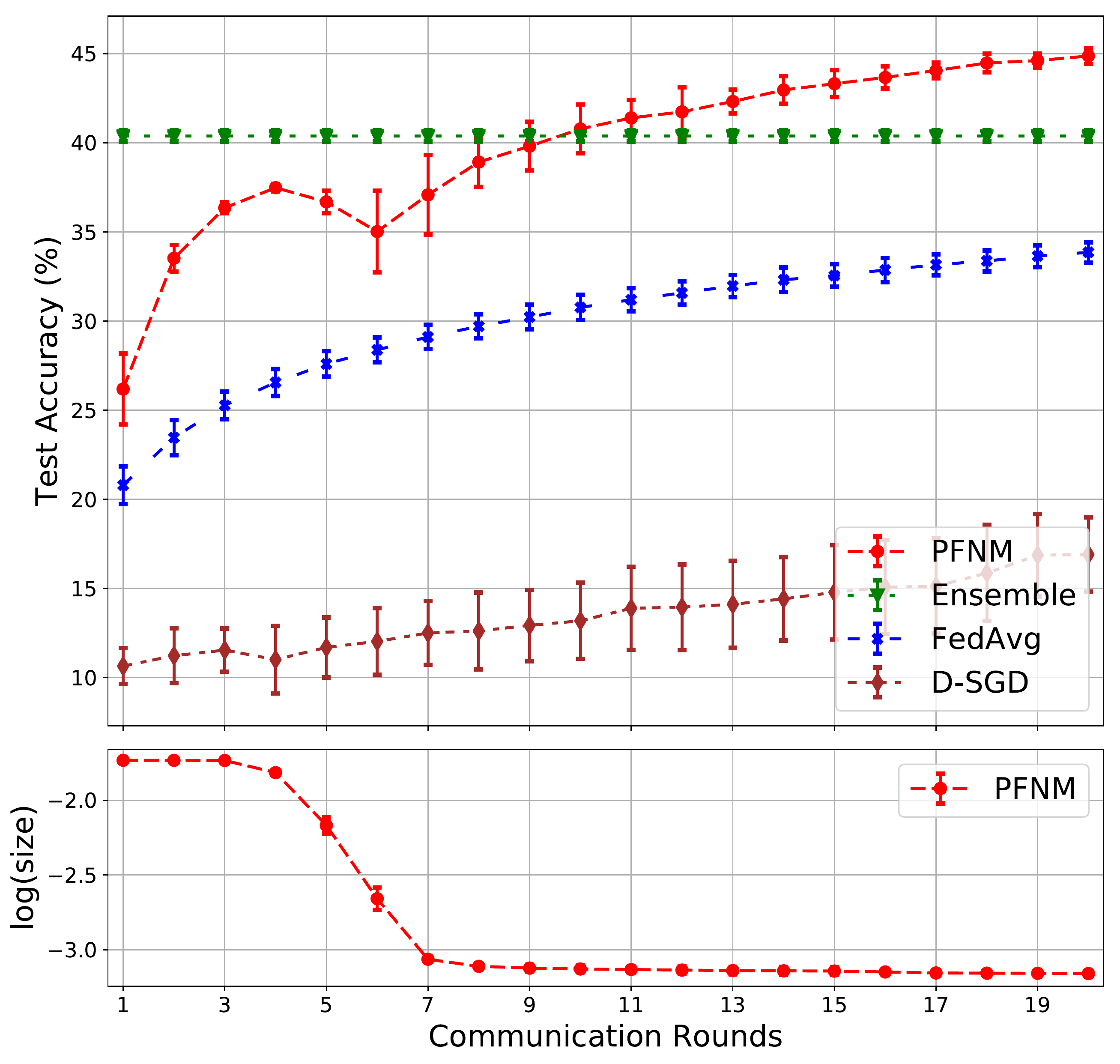}
\caption{CIFAR homogeneous}
\label{fig:cifar_2_homo_comm}
\end{subfigure}
\begin{subfigure}{.245\textwidth}
  \centering
  \captionsetup{justification=centering}
\includegraphics[width=\linewidth]{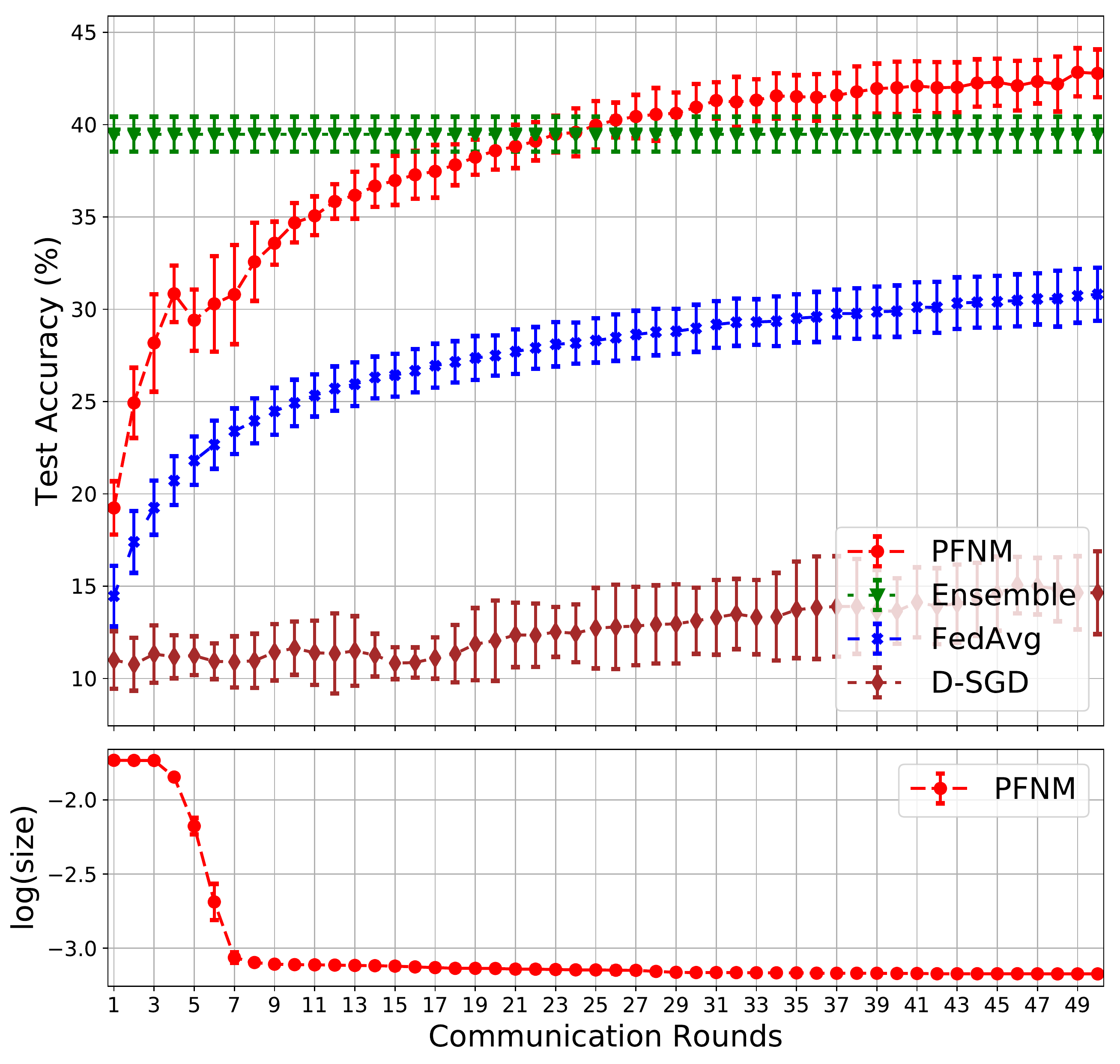}
\caption{CIFAR heterogeneous}
\label{fig:cifar_2_hetero_comm}
\end{subfigure}
\caption{\textbf{Federated learning with communications}. Test accuracy and normalized model size as a function of number of communication rounds for $J=25$ batches for one (TOP) and two layer (BOTTOM) neural networks. PFNM consistently outperforms strong competitors.}
\label{fig:comm_1}
\vskip -0.1in
\end{figure*}

\paragraph{Learning with limited communication}
While in some scenarios limiting communication to a single communication round may be a hard constraint, we also consider situations, that frequently arise in practice, where a limited amount of communication is permissible. To this end, we investigate federated learning with $J=25$ batches and up to twenty communications when the data has a homogeneous partition and up to fifty communications under a heterogeneous partition. We compare PFNM, using the communication procedure from Section \ref{subsec:communication} ($\sigma=\sigma_0=\gamma_0=1$ across experiments) to federated averaging and the distributed optimization approach, downpour SGD (D-SGD) of \citet{dean2012large}. In this limited communication setting, the ensembles can be outperformed by many distributed learning algorithms provided a large enough communication budget. An interesting metric then is the number of communications rounds required to outperform ensembles. 

We report results with both one and two layer neural networks in Figure~\ref{fig:comm_1}. In either case, we use a hundered neurons per layer. PFNM outperforms ensembles in all scenarios given sufficient communications. Moreover, in all experiments, PFNM requires significantly fewer communication rounds than both federated averaging and D-SGD to achieve a given performance level. In addition to improved performance, additional rounds of communication allow PFNM to shrink the size of the global model as demonstrated in the figure. In \Cref{fig:mnist_1_homo_comm,fig:mnist_1_hetero_comm,fig:mnist_2_homo_comm,fig:mnist_2_hetero_comm,fig:cifar_1_homo_comm,fig:cifar_1_hetero_comm} we note steady improvement in accuracy and reduction in the global model size. In CIFAR10 experiments, the two layer PFNM network's performance temporarily drops, which corresponds to a sharp reduction in the size of the global network. See Figures \ref{fig:cifar_2_homo_comm} and \ref{fig:cifar_2_hetero_comm}.  

\section{Discussion}
\label{sec:discussion}
In this work, we have developed methods for federated learning of neural networks, and empirically demonstrated their favorable properties. Our methods are particularly effective at learning compressed federated networks from pre-trained local networks and with a modest communication budget can outperform state-of-the-art algorithms for federated learning of neural networks.  
In future work, we plan to explore more sophisticated ways of combining local networks especially in the regime where each local network has very few training instances. Our current matching approach is completely unsupervised -- incorporating some form of supervision may help further improve the performance of the global network, especially when the local networks are of poor quality.
Finally, it is of interest to extend our modeling framework to other architectures such as Convolutional Neural Networks (CNNs) and Recurrent Neural Networks (RNNs). 
The permutation invariance necessitating matching inference also arises in CNNs since any permutation of the filters results in the same output, however additional bookkeeping is needed due to the pooling operations.

\clearpage
\bibliography{MY_ref}
\bibliographystyle{icml2019}

\clearpage
\appendix

\section{Single Hidden Layer Inference}\label{supp:SinglLayer}
The goal of maximum a posteriori (MAP) estimation is to maximize posterior probability of the latent variables: global atoms $\{\rvtheta_i\}_{i=1}^\infty$ and assignments of observed neural network weight estimates to global atoms $\{\mB^j\}_{j=1}^J$, given estimates of the batch weights $\{\rvv_{jl}\text{ for }l=1,\ldots,L_j\}_{j=1}^J$:
\begin{align}
    \label{eq:supp:map}
    \argmax\limits_{\{\rvtheta_i\},\{\mB_j\}}P(\{\rvtheta_i\},&\{\mB^j\}|\{\rvv_{jl}\}) \propto\\&\nonumber P(\{\rvv_{jl}\}|\{\rvtheta_i\},\{\mB^j\})P(\{\mB^j\})P(\{\rvtheta_i\}).
\end{align}
\paragraph{MAP estimates given matching.} First we note that given $\{\mB^j\}$ it is straightforward to find MAP estimates of $\{\rvtheta_i\}$ based on Gaussian-Gaussian conjugacy:
\begin{equation}
\label{eq:supp:gaus_map_general}
\hat \rvtheta_i = \frac{\sum_{j,l}\emB^j_{i,l}\rvv_{jl}/\sigma_{j}^2}{1/\sigma_0^2 + \sum_{j,l}\emB^j_{i,l}/\sigma_{j}^2}\text{ for }i=1,\ldots,L,
\end{equation}
where $L = \max\{i:\emB^j_{i,l} = 1\text{ for }l=1,\ldots,L_j,\ j=1,\ldots,J\}$ is the number of active global atoms, which is an (unknown) latent random variable identified by $\{\mB^j\}$. For simplicity we assume $\mSigma_0 = \mI\sigma^2_0$, $\mSigma_j = \mI\sigma^2_j$ and $\vmu_0=0$. 

\paragraph{Inference of atom assignments (Proposition \ref{Prop:CostMatrix}).} We can now cast optimization corresponding to \eqref{eq:supp:map} with respect to only $\{\mB^j\}_{j=1}^J$. Taking natural logarithm we obtain:
\begin{align}
\nonumber
-\frac{1}{2}\sum_i&\left(\frac{\|\hat \rvtheta_i\|^2}{\sigma_0^2} + D\log(2\pi\sigma_0^2) +  \sum_{j,l}\emB^j_{i,l}\frac{\|\rvv_{jl}-\hat \rvtheta_i\|^2}{\sigma_{j}^2}\right)\\\label{eq:supp:objective_multi_batch} & \qquad \qquad\qquad + \log(P(\{\mB^j\}).
\end{align}
We now simplify the first term of \eqref{eq:supp:objective_multi_batch} (in this and subsequent derivations we use $\cong$ to say that two objective functions are equivalent up to terms independent of the variables of interest):

\begin{equation}
\label{eq:supp:objective_multi_batch_simplified}
\begin{split}
 -&\frac{1}{2}\sum_i\left(\frac{\|\hat \rvtheta_i\|^2}{\sigma_0^2} + D\log(2\pi\sigma_0^2) +  \sum_{j,l}\emB^j_{i,l}\frac{\|\rvv_{jl}-\hat \rvtheta_i\|^2}{\sigma_{j}^2}\right) \\
= & -\frac{1}{2}\sum_i\left(\frac{\langle \hat \rvtheta_i,\hat \rvtheta_i\rangle}{\sigma_0^2} + D\log(2\pi\sigma_0^2)\right.\\ &\hspace{40pt}+\left. \sum_{j,l}\emB^j_{i,l}\frac{\langle \rvv_{jl},\rvv_{jl}\rangle - 2\langle \rvv_{jl},\hat \rvtheta_i \rangle + \langle \hat \rvtheta_i, \hat \rvtheta_i \rangle}{\sigma^2_{jl}}\right)\\
\cong & -\frac{1}{2}\sum_i\left(\langle \hat \rvtheta_i,\hat \rvtheta_i\rangle\left(\frac{1}{\sigma_0^2} + \sum_{j,l}\frac{\emB^j_{i,l}}{\sigma_{j}^2}\right) + D\log(2\pi\sigma_0^2) \right.\\&\hspace{130pt}-\left. 2 \langle \hat \rvtheta_i, \sum_{j,l}\emB^j_{i,l}\frac{\rvv_{jl}}{\sigma_{j}^2}\rangle\right)\\
= & \frac{1}{2}\sum_i\left(\langle \hat \rvtheta_i,\hat \rvtheta_i\rangle\left(\frac{1}{\sigma_0^2} + \sum_{j,l}\frac{\emB^j_{i,l}}{\sigma_{j}^2}\right) - D\log(2\pi\sigma_0^2)\right) \\
= & \frac{1}{2}\sum_i \left( \frac{\|\sum_{j,l}\emB^j_{i,l}\rvv_{jl}/\sigma_{j}^2\|^2}{1/\sigma_0^2+\sum_{j,l}\emB^j_{i,l}/\sigma_{j}^2} - D\log(2\pi\sigma_0^2) \right).
\end{split}
\end{equation}
We consider an iterative optimization approach: fixing all but one $\mB^j$ we find the corresponding optimal assignment, then pick a new $j$ at random and repeat until convergence. We define notation $-j$ to denote ``all but $j$'', and let $L_{-j} = \max\{i: B^{-j}_{i,l} = 1\}$ denote number of active global weights outside of group $j$. We partition \eqref{eq:supp:objective_multi_batch_simplified} between $i=1,\ldots,L_{-j}$ and $i=L_{-j}+1,\ldots,L_{-j}+L_j$, and since we are solving for $\mB^j$, we subtract terms independent of $\mB^j$:
\begin{align}
\nonumber
     \sum_i &\left( \frac{\|\sum_{j,l}\emB^j_{i,l}\rvv_{jl}/\sigma_{j}^2\|^2}{1/\sigma_0^2+\sum_{j,l}\emB^j_{i,l}/\sigma_{j}^2} - D\log(2\pi\sigma_0^2) \right) \\\nonumber
    &\cong  \sum_{i=1}^{L_{-j}}\left( \frac{\|\sum_l\emB^j_{i,l}\rvv_{jl}/\sigma_{j}^2 + \sum_{-j,l}\emB^j_{i,l}\rvv_{jl}/\sigma_{j}^2\|^2}{1/\sigma_0^2 + \sum_l\emB^j_{i,l}/\sigma_{j}^2 + \sum_{-j,l}\emB^j_{i,l}/\sigma_{j}^2}\right. \\\nonumber &\qquad\qquad\qquad\qquad\left.- \frac{\|\sum_{-j,l}\emB^j_{i,l}\rvv_{jl}/\sigma_{j}^2\|^2}{1/\sigma_0^2+\sum_{-j,l}\emB^j_{i,l}/\sigma_{j}^2}\right) \\
    &\quad+  \sum_{i=L_{-j}+1}^{L_{-j}+L_j} \left(\frac{\|\sum_l\emB^j_{i,l}\rvv_{jl}/\sigma_{j}^2\|^2}{1/\sigma_0^2 + \sum_l\emB^j_{i,l}/\sigma_{j}^2}\right).\label{eq:objective_multi_batch_partitioned}
\end{align}
Now observe that $\sum_l\emB^j_{i,l}\in\{0,1\}$, i.e. it is 1 if some neuron from batch $j$ is matched to global neuron $i$ and 0 otherwise. Due to this we can rewrite \eqref{eq:objective_multi_batch_partitioned} as a linear sum assignment problem:
\begin{align}
\nonumber
 \sum_{i=1}^{L_{-j}}&\sum_{l=1}^{L_j} \emB^j_{i,l}\left(\frac{\|\rvv_{jl}/\sigma_{j}^2 + \sum_{-j,l}\emB^j_{i,l}\rvv_{jl}/\sigma_{j}^2\|^2}{1/\sigma_0^2 + 1/\sigma_{j}^2 + \sum_{-j,l}\emB^j_{i,l}/\sigma_{j}^2} \right.\\\nonumber&\hspace{100pt}-\left. \frac{\|\sum_{-j,l}\emB^j_{i,l}\rvv_{jl}/\sigma_{j}^2\|^2}{1/\sigma_0^2+\sum_{-j,l}\emB^j_{i,l}/\sigma_{j}^2}\right)\\
&+  \sum_{i=L_{-j} + 1}^{L_{-j} + L_j}\sum_{l=1}^{L_j} \emB^j_{i,l}\left(\frac{\|\rvv_{jl}/\sigma_{j}^2\|^2}{1/\sigma_0^2 + 1/\sigma_{j}^2}\right).
\label{eq:supp:objective_j_hungarian_param}
\end{align}
Now we consider second term of \eqref{eq:supp:objective_multi_batch}:
\begin{equation*}
\log P(\{\mB^j\}) = \log P(\mB^j|\mB^{-j}) + \log P(\mB^{-j}).
\end{equation*}
First, because we are optimizing for $\mB^j$, we can ignore $\log P(\mB^{-j})$. Second, due to exchangeability of batches (i.e. customers of the IBP), we can always consider $\mB^j$ to be the last batch (i.e. last customer of the IBP). Let $m^{-j}_i = \sum_{-j,l}\emB^j_{i,l}$ denote number of times batch weights were assigned to global atom $i$ outside of group $j$. We now obtain the following:
\begin{equation}
\label{eq:objective_nonparam_partitioned}
\begin{split}
&\log P(\mB^j|\mB^{-j}) \cong \\
& \sum_{i=1}^{L_{-j}} \left(\mspace{-5mu}\left(\mspace{-2mu}\sum_{l=1}^{L_j}\mspace{-2mu}\emB^j_{i,l}\mspace{-2mu}\right)\mspace{-2mu}\log\mspace{-2mu}\frac{m^{-j}_i}{J} \mspace{-2mu}+\mspace{-2mu} \left(\mspace{-2mu}1\mspace{-2mu}-\mspace{-2mu}\sum_{l=1}^{L_j}\emB^j_{i,l}\right)\mspace{-2mu}\log\frac{J - m^{-j}_i}{J}\right)\\
&-  \log\left(\sum_{i=L_{-j}+1}^{L_{-j}+L_j}\sum_{l=1}^{L_j}\emB^j_{i,l}\right) + \left(\sum_{i=L_{-j}+1}^{L_{-j}+L_j}\sum_{l=1}^{L_j}\emB^j_{i,l}\right)\log\frac{\gamma_0}{J}.
\end{split}
\end{equation}
We now rearrange \eqref{eq:objective_nonparam_partitioned} as linear sum assignment problem:
\begin{align}
\label{eq:supp:objective_j_hungarian_nonparam}
\sum_{i=1}^{L_{-j}}\sum_{l=1}^{L_j} &\emB^j_{i,l}\log\frac{m^{-j}_i}{J - m^{-j}_i} \\\nonumber&+ \sum_{i=L_{-j} + 1}^{L_{-j} + L_j}\sum_{l=1}^{L_j} \emB^j_{i,l}\left(\log\frac{\gamma_0}{J} - \log(i-L_{-j})\right).
\end{align}
Combining \eqref{eq:supp:objective_j_hungarian_param} and \eqref{eq:supp:objective_j_hungarian_nonparam} we arrive at the cost specification for finding $\mB^j$ as \emph{minimizer} of $\sum_i\sum_l\emB^j_{i,l}\emC^j_{i,l}$, where:
\begin{align}
\label{eq:supp:cost_J}
&\emC^j_{i,l} = \\\nonumber
&- \begin{cases} \begin{aligned}[b]\frac{\left\|\frac{\rvv_{jl}}{\sigma_{j}^2} + \sum\limits_{-j,l}\emB^j_{i,l}\frac{\rvv_{jl}}{\sigma_{j}^2}\right\|^2}{\frac{1}{\sigma_0^2} + \frac{1}{\sigma_{j}^2} + \sum_{-j,l}\emB^j_{i,l}/\sigma_{j}^2} &- \frac{\left\|\sum\limits_{-j,l}\emB^j_{i,l}\frac{\rvv_{jl}}{\sigma_{j}^2}\right\|^2}{\frac{1}{\sigma_0^2}+\sum_{-j,l}\emB^j_{i,l}/\sigma_{j}^2}\\ + 2\log&\frac{m^{-j}_i}{J - m^{-j}_i},
\end{aligned} &\hspace{-35pt} i \leq L_{-j} \\
\frac{\left\|\frac{\rvv_{jl}}{\sigma_{j}^2}\right\|^2}{\frac{1}{\sigma_0^2} + \frac{1}{\sigma_{j}^2}} - 2\log\frac{i-L_{-j}}{\gamma_0/J} , & \hspace{-90pt} L_{-j} < i \leq L_{-j} + L_j.
\end{cases}
\end{align}
This completes the proof of Proposition \ref{Prop:MAP}.

\section{Multilayer Inference Details}
\label{supp:multilayer}
\begin{figure}[h!]
    \centering
    \includegraphics[width=3in]{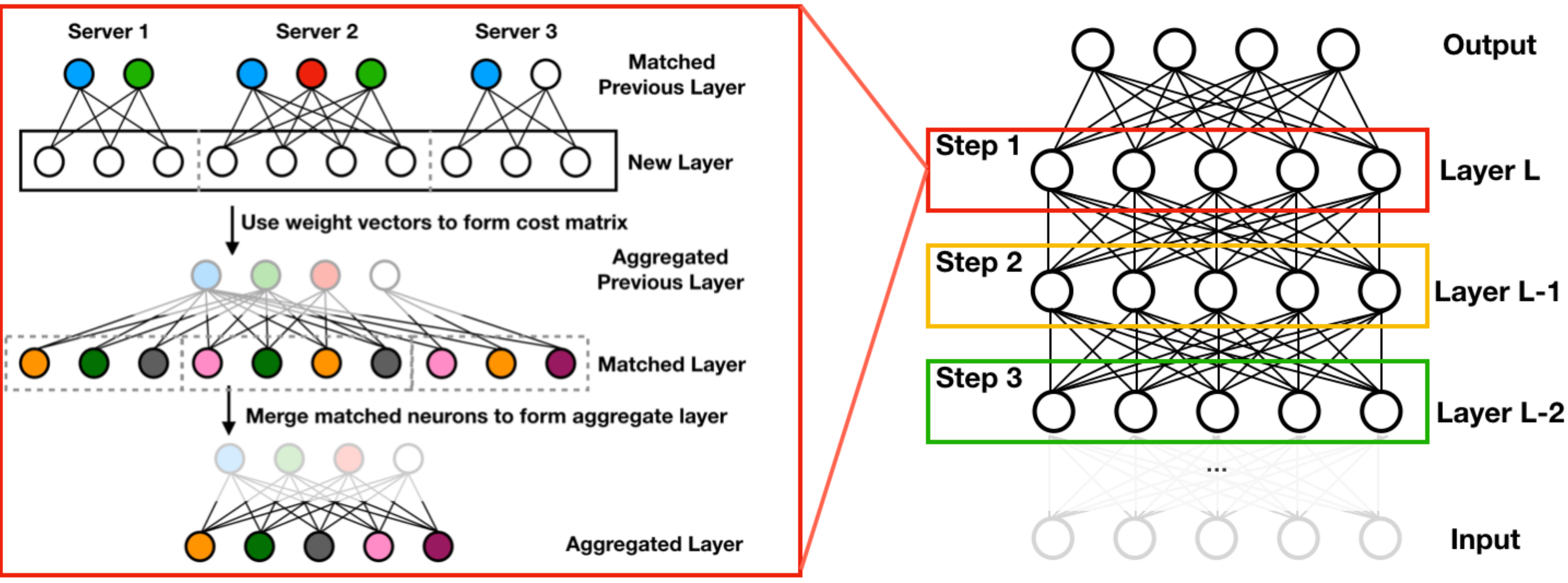}
    \caption{Probabilistic Federated Neural Matching algorithm showing matching of three multilayer MLPs. Nodes in the graphs indicate neurons, neurons of the same color have been matched. On the left, the individual layer matching approach is shown, consisting of using the matching assignments of the next highest layer to convert the neurons in each of the $J$ batches to weight vectors referencing the global previous layer. These weight vectors are then used to form a cost matrix, which the Hungarian algorithm then uses to do the matching. Finally, the matched neurons are then aggregated and averaged to form the new layer of the global model. As shown on the right, in the multilayer setting the resulting global layer is then used to match the next lower layer, etc. until the bottom hidden layer is reached (Steps 1, 2, 3,... in order).   }
    \label{Fig:Diagram}
\end{figure}

Figure \ref{Fig:Diagram} illustrates the overall multilayer inference procedure visually, and Algorithm \ref{alg:MultiLayer} provides the details. 

\begin{algorithm}[h]
\caption{Multilayer PFNM}
\begin{algorithmic}[1]
\STATE $L^{C+1} \gets $ number of outputs
\STATE\# Top down iteration through layers
\FOR{layers $c = C, C-1, \dots, 2$} 
\STATE Collect hidden layer $c$ from the $J$ batches and form $\rvv_{jl}^c$. 
\STATE Call Single Layer Neural Matching algorithm with output dimension $L^{c+1}$ and input dimension 0 (since we do not use the weights connecting to lower layers here).
\STATE Form global neuron layer $c$ from output of the single layer matching.
\STATE $L^{c} \gets \mathrm{card}(\cup_{j=1}^J \mathcal{T}_j^c)$ (greedy approach).
\ENDFOR
\STATE \# Match bottom layer using weights connecting to both the input and the layer above.
\STATE Call Single Layer Neural Matching algorithm with output dimension $L^{2}$ and input dimension equal to the number of inputs.
\STATE Return global assignments and form global mutltilayer model.
\end{algorithmic}
\label{alg:MultiLayer}
\end{algorithm}

\section{Complexity Analysis}
\label{supp:complexity}
In this section we present a brief discussion of the complexity of our algorithms. The worst case complexity per layer is achieved when no neurons are matched and is equal to $\mathcal{O}(D(J L_j)^2)$ for building the cost matrix and $\mathcal{O}((J L_j)^3)$ for running the Hungarian algorithm, where $L_j$ is the number of neurons per batch (here for simplicity we assume that each batch has same number of neurons) and $J$ is the number of batches. The best case complexity per layer (i.e. when all neurons are matched) is $\mathcal{O}(D L_j^2 + L_j^3)$, also note that complexity is independent of the data size. In practice the complexity is closer to the best case since global model size is moderate (i.e. $L \ll \sum_j L_j$). Actual timings with our code for the experiments in Figure \ref{fig:J100} are as follows - 40sec for Fig. \ref{fig:J100_mnist_homo_acc}, \ref{fig:J100_mnist_homo_shape} at $J=30$ groups; 500sec for Fig. \ref{fig:J100_mnist_hetero_acc}, \ref{fig:J100_mnist_hetero_shape} at $J=30$ (the $D L_j^2$ term is dominating as CIFAR10 dimension is much higher than MNIST); 60sec for Fig. \ref{fig:C_mnist_homo}, \ref{fig:C_mnist_hetero} ($J=10$) at $C=6$ layers; 150sec for Fig. \ref{fig:C_cifar_homo}, \ref{fig:C_cifar_hetero} ($J=10$) at $C=6$. The computations were done using 2 CPU cores and 4GB memory on a machine with 3.0 GHz core speed. We note that (i) this computation only needs to be performed once (ii) the cost matrix construction which appears to be dominating can be trivially sped up using GPUs (iii) recent work demonstrates impressive large scale running times for the Hungarian algorithm using GPUs \citep{date2016gpu}.

\begin{figure*}[t!]
\centering
\begin{subfigure}{.45\textwidth}
  \centering
  \captionsetup{justification=centering}
\includegraphics[width=\linewidth]{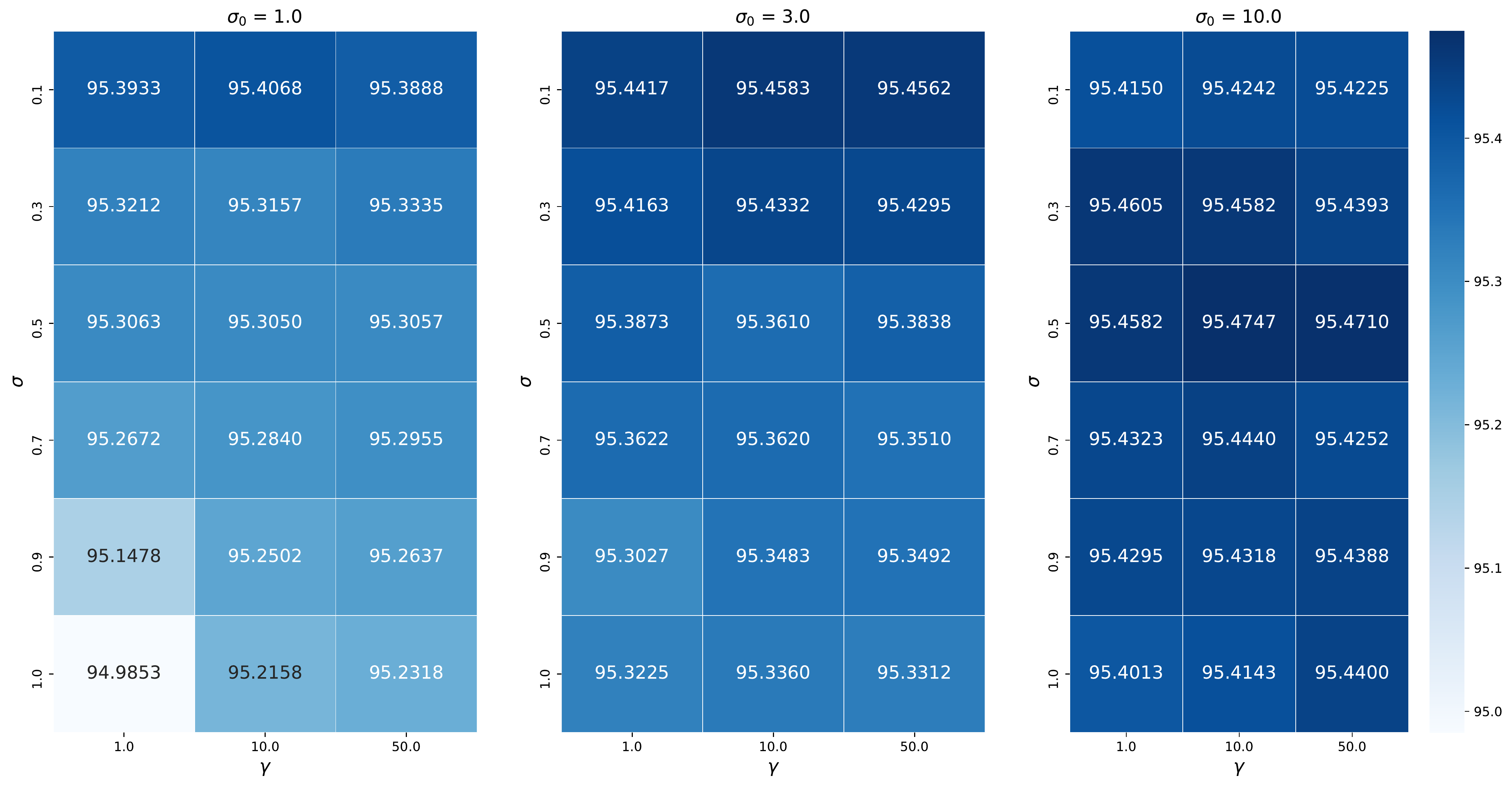}
\caption{MNIST homogeneous}
\label{fig:J_mnist_homo}
\end{subfigure}
\begin{subfigure}{.45\textwidth}
  \centering
  \captionsetup{justification=centering}
\includegraphics[width=\linewidth]{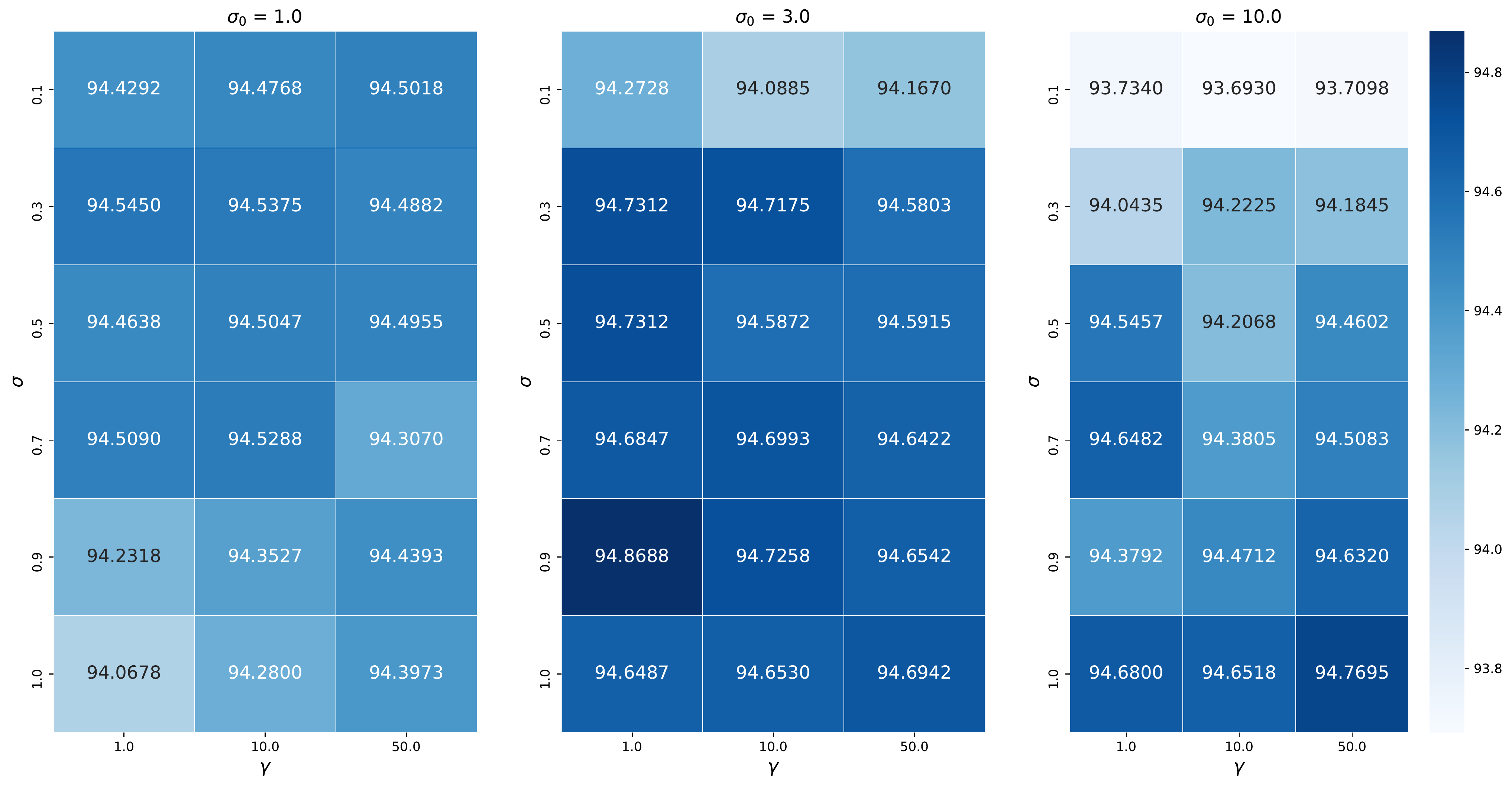}
\caption{MNIST heterogeneous}
\label{fig:J_mnist_hetero}
\end{subfigure}
\begin{subfigure}{.45\textwidth}
  \centering
  \captionsetup{justification=centering}
\includegraphics[width=\linewidth]{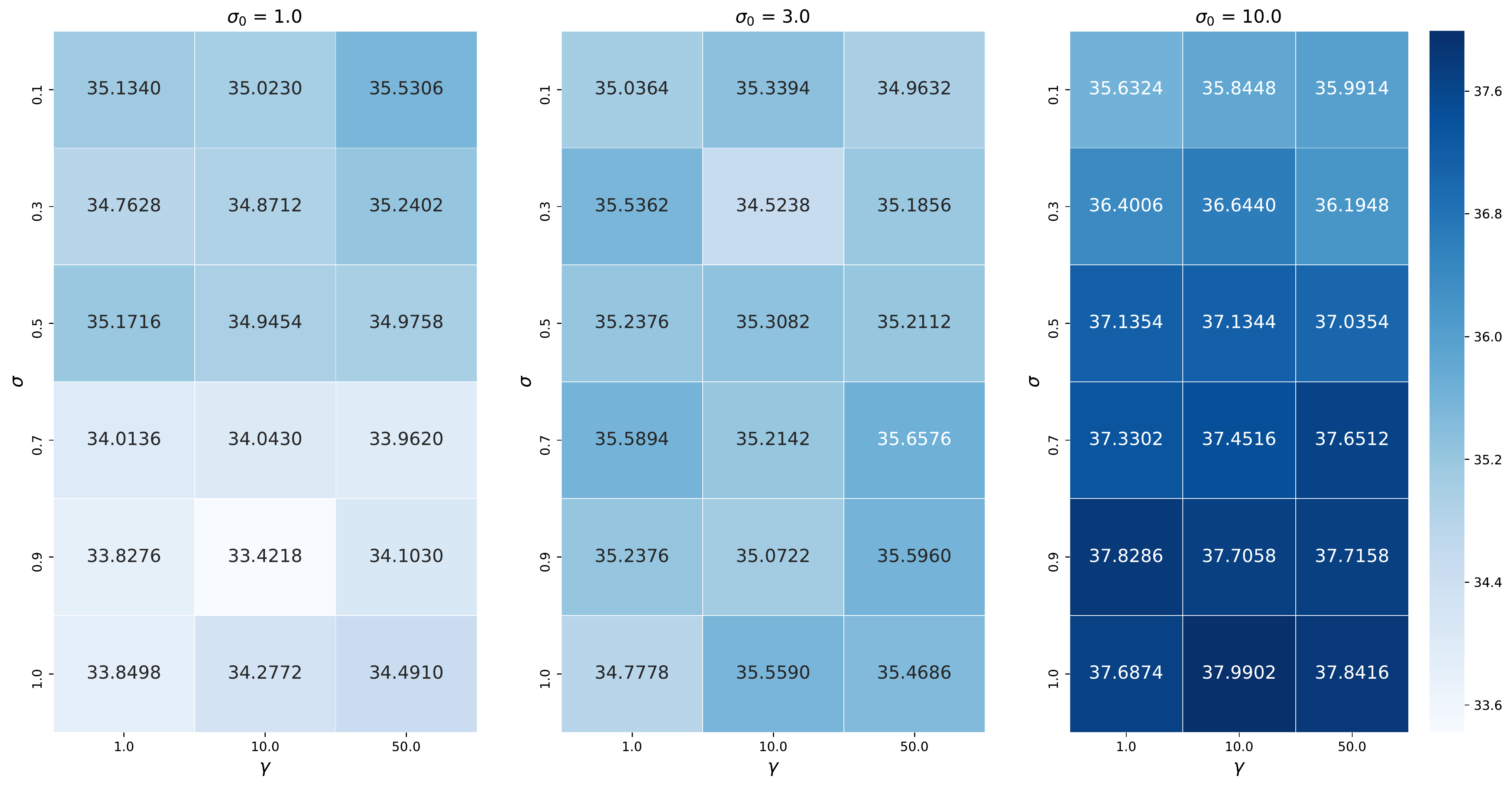}
\caption{CIFAR homogeneous}
\label{fig:J_cifar_homo}
\end{subfigure}
\begin{subfigure}{.45\textwidth}
  \centering
  \captionsetup{justification=centering}
\includegraphics[width=\linewidth]{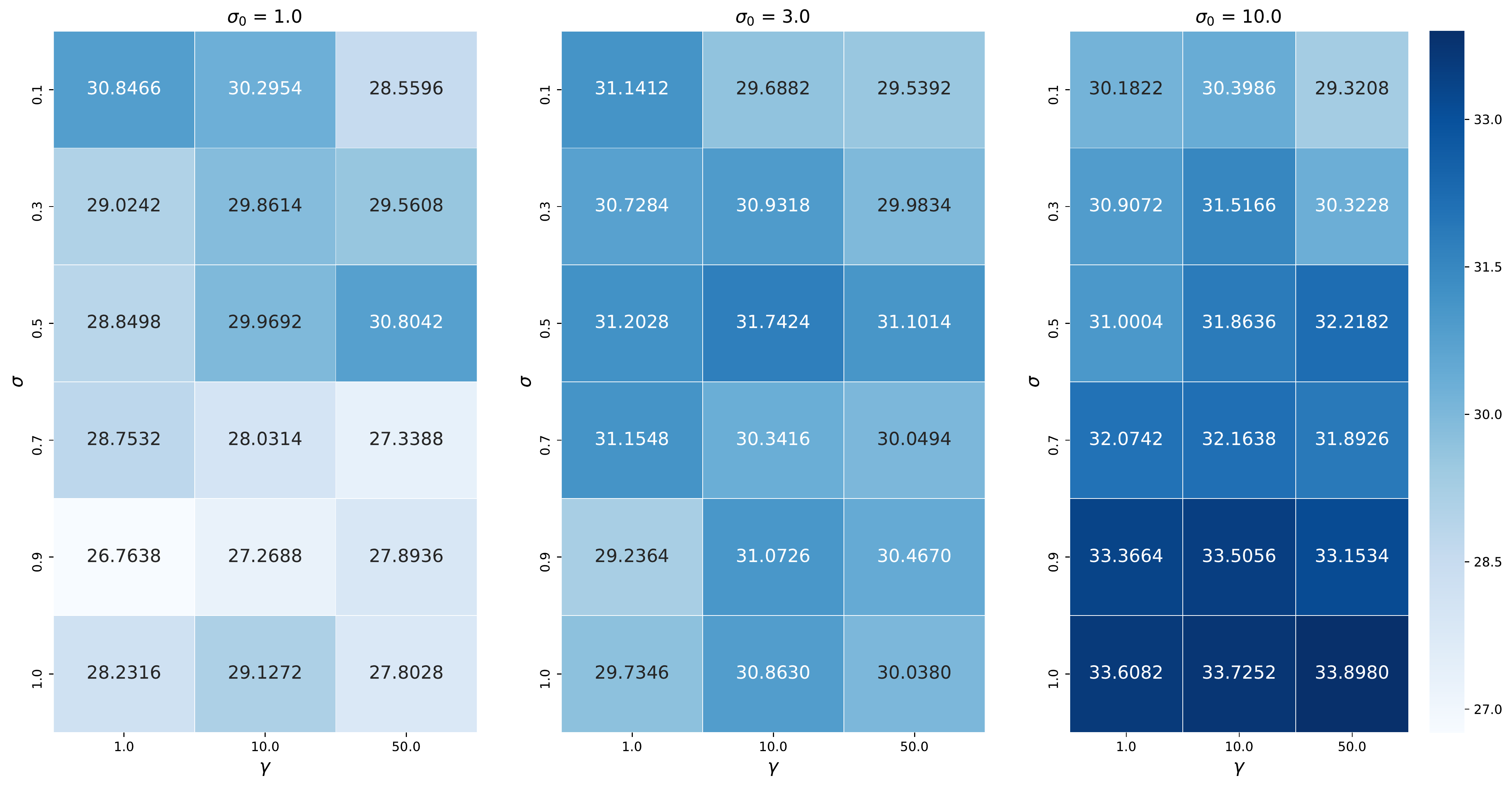}
\caption{CIFAR heterogeneous}
\label{fig:J_cifar_hetero}
\end{subfigure}
\caption{Parameter sensitivity analysis for $J = 25$.}
\label{fig:sens}
\end{figure*}

\begin{figure}[h!]
\centering
\begin{subfigure}{.23\textwidth}
  \centering
  \captionsetup{justification=centering}
\includegraphics[width=\linewidth]{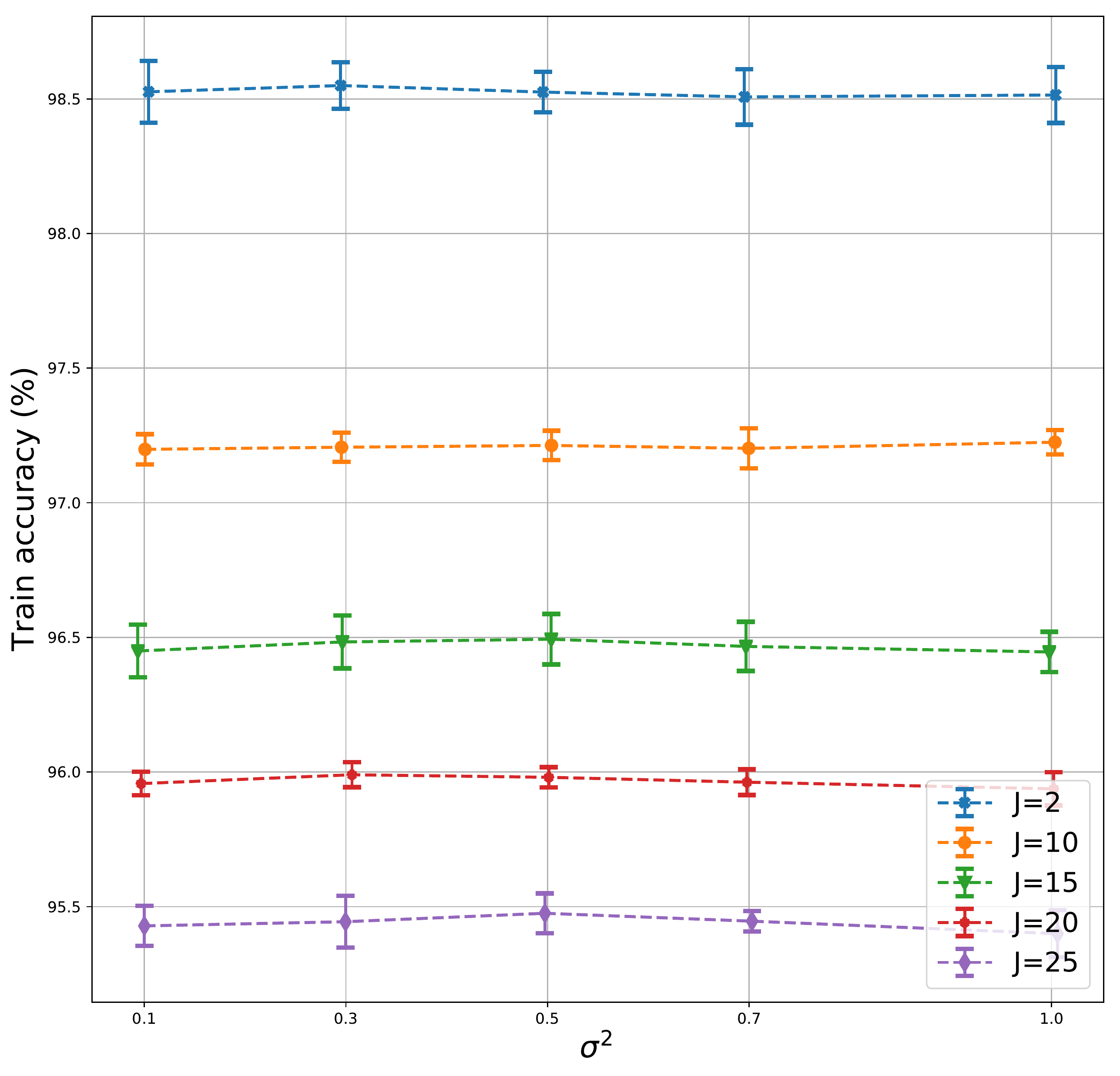}
\caption{MNIST homogeneous}
\label{fig:J_sigma_mnist_homo}
\end{subfigure}
\begin{subfigure}{.23\textwidth}
  \centering
  \captionsetup{justification=centering}
\includegraphics[width=\linewidth]{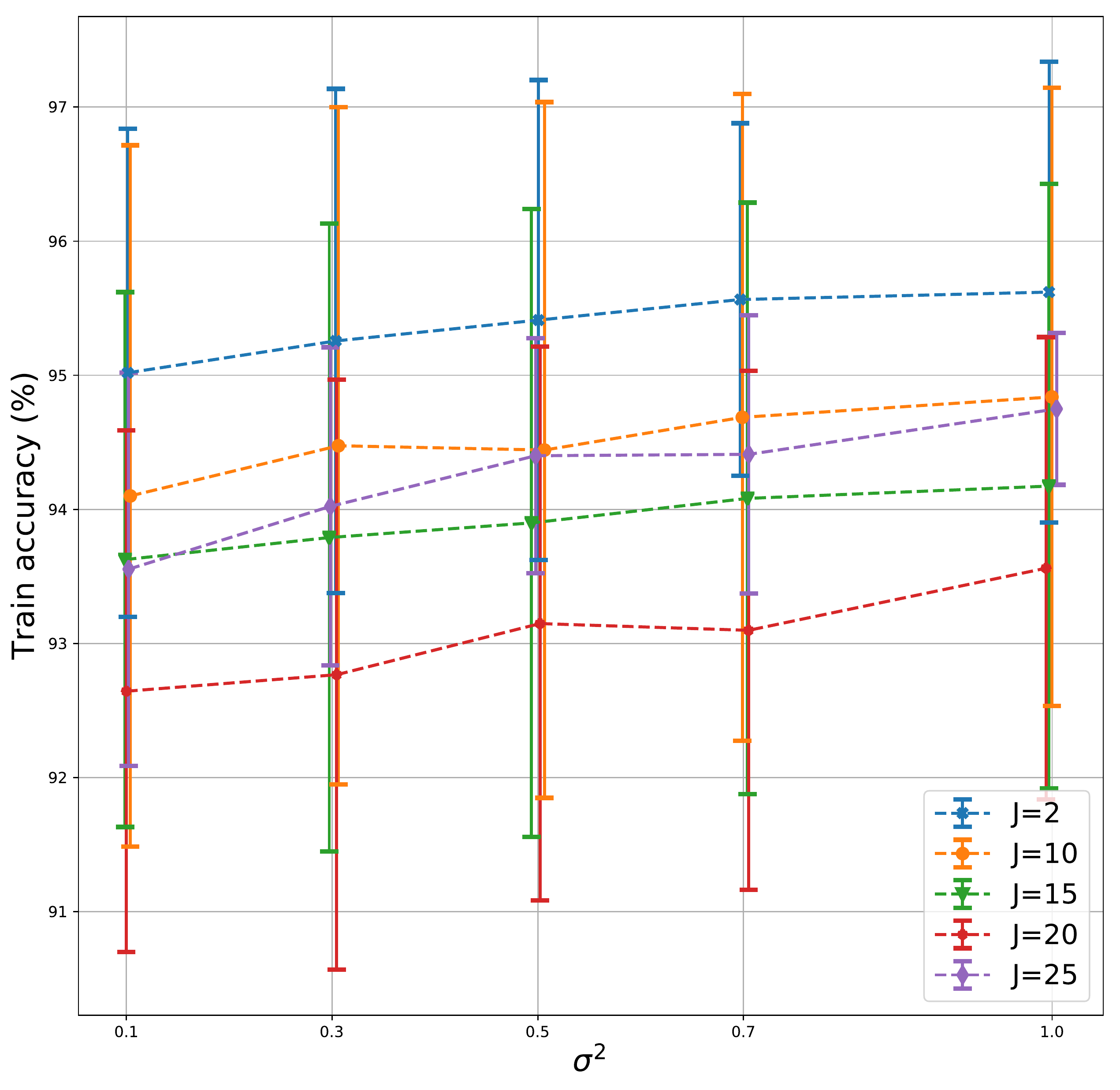}
\caption{MNIST heterogeneous}
\label{fig:J_sigma_mnist_hetero}
\end{subfigure}
\\
\begin{subfigure}{.23\textwidth}
  \centering
  \captionsetup{justification=centering}
\includegraphics[width=\linewidth]{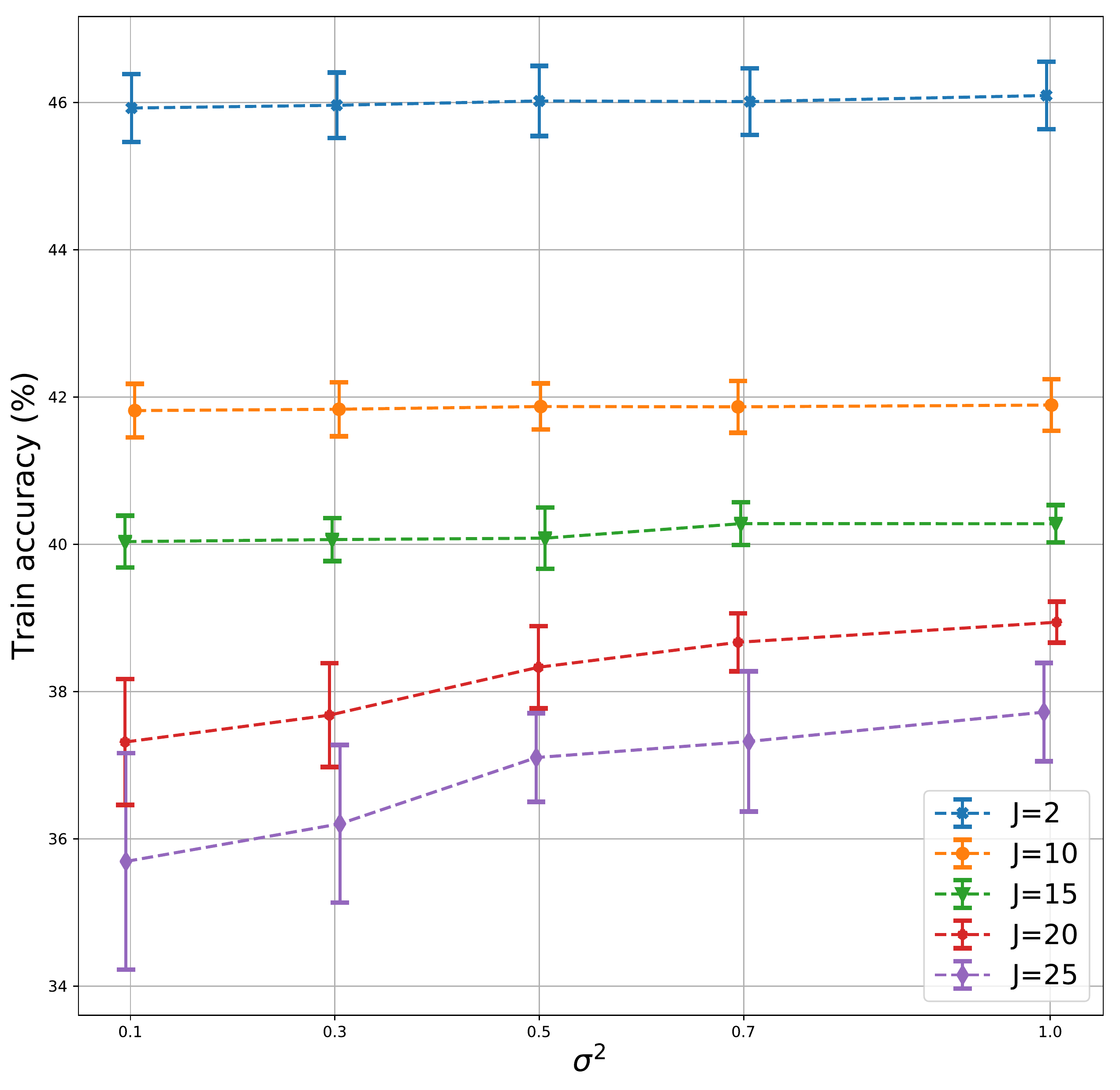}
\caption{CIFAR homogeneous}
\label{fig:J_sigma_cifar_homo}
\end{subfigure}
\begin{subfigure}{.23\textwidth}
  \centering
  \captionsetup{justification=centering}
\includegraphics[width=\linewidth]{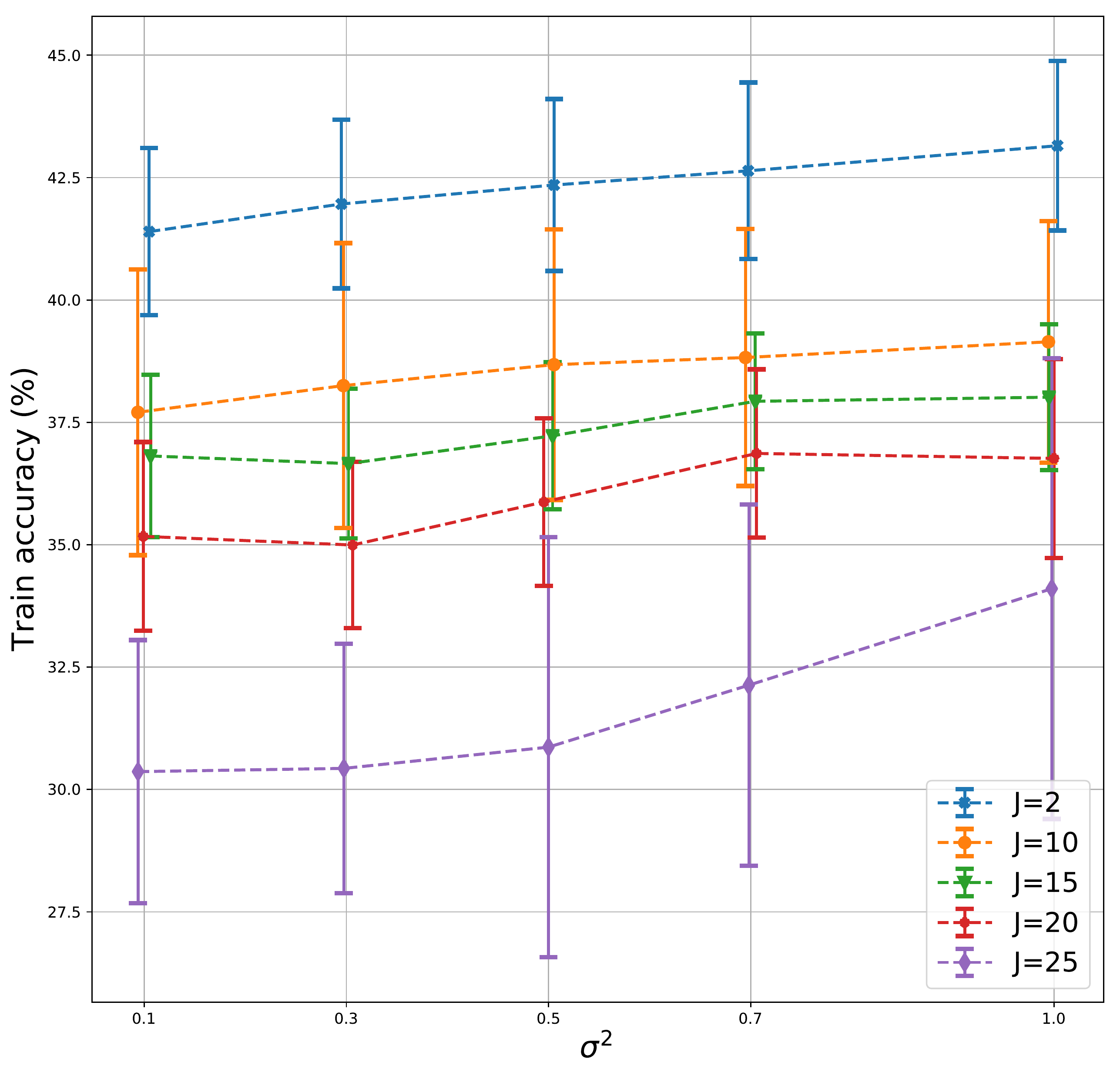}
\caption{CIFAR heterogeneous}
\label{fig:J_sigma_cifar_hetero}
\end{subfigure}
\caption{Sensitivity analysis of $\sigma^2$ for fixed $\sigma_0^2=10$ and $\gamma_0=1$ for varying $J$.}
\label{fig:J_sigma}
\end{figure}

\section{Experimental Details and Additional Results}
\label{supp:exp_details}

\paragraph{Data partitioning.} In the federated learning setup, we analyze data from multiple sources, which we call batches. Data on the batches does not overlap and may have different distributions. To simulate federated learning scenario we consider two partition strategies of MNIST and CIFAR-10. For each pair of partition strategy and dataset we run 10 trials to obtain mean accuracies and standard deviations. The easier case is homogeneous partitioning, i.e. when class distributions on batches are approximately equal as well as batch sizes. To generate homogeneous partitioning with $J$ batches we split examples for each of the classes into $J$ approximately equal parts to form $J$ batches. In the heterogeneous case, batches are allowed to have highly imbalanced class distributions as well as highly variable sizes. To simulate heterogeneous partition, for each class $k$, we sample $\rvp_k \sim \text{Dir}_J(0.5)$ and allocate $\rvp_{k,j}$ proportion of instances of class $k$ of the complete dataset to batch $j$. Note that due to small concentration parameter, 0.5, of the Dirichlet distribution, some batches may entirely miss examples of a subset of classes.

\paragraph{Batch networks training.}
Our modeling framework and ensemble related methods operate on collection of weights of neural networks from all batches. Any optimization procedure and software can be used locally on batches for training neural networks. We used PyTorch \citep{paszke2017automatic} to implement the networks and train these using the  AMSGrad optimizer \citep{amsgrad} with default parameters unless otherwise specified. For reproducibility we summarize all parameter settings in Table \ref{table:batch_parameters}.
\begin{table}[h]
\caption{Parameter settings for batch neural networks training}
\label{table:batch_parameters}
\begin{center}
\begin{tabular}{lll}
 & MNIST & CIFAR-10
\\ \midrule \\
Neurons per layer & 100 & 100 \\
Learning rate   & 0.01 & 0.001 \\
$L_2$ regularization & $10^{-6}$ & $10^{-5}$ \\
Minibatch size       & 32 & 32 \\
Epochs              & 10 & 10 \\
Weights initialization & $\mathcal{N}(0,0.01)$ & $\mathcal{N}(0,0.01)$ \\
Bias initialization & 0.1 & 0.1
\end{tabular}
\end{center}
\end{table}

\subsection{Parameter Settings for the Baselines}
We first formally define the ensemble procedure. Let $\hat y_j \in \Delta^{K-1}$ denote the probability distribution over the $K$ classes output by neural network trained on data from batch $j$ for some test input $x$. Then ensemble prediction is $\argmax\limits_{k}\frac{1}{J}\sum_{j=1}^J \hat y_{j,k}$. In our experiments, we train each individual network on the specific batch dataset using the parameters listed in Table \ref{table:batch_parameters}, and then compute the performance using the ensemble aggregation technique.


For the downpour SGD \citep{dean2012large} we used PyTorch, SGD optimizer and parameter settings as in Table \ref{table:batch_parameters} for the local learners. The master neural network was optimized with Adam and the same initial learning rate as in the Table \ref{table:batch_parameters}. The local learners communicated the accumulated gradients back to the master network after every mini-batch update. This translates to the setting of \citet{dean2012large} with parameters $n_{push} = n_{fetch} = 1$. Note that with this approach the global network and networks for each of the batches are constrained to have identical number of neurons per layer, which is 100 in our experiments.

For Federated Averaging \citep{mcmahan2017communication}, we use SGD optimizer for learning the local networks with the rest of the parameters as defined in Table \ref{table:batch_parameters}. We initialize all the local networks with the same seed, and train these networks for 10 epochs initially and for 5 epochs after the first communication round. At each communication round, we utilize all the local networks ($C=1$) for the central model update. 

\subsection{Parameter Settings for Matching with Additional Communications}

For neural matching with additional communications, we train the local networks for 10 epochs for the first communication round, and 5 epochs thereafter. All the other parameters are as mentioned in Table \ref{table:batch_parameters}. The local networks are trained using AMSGrad optimizer \citep{amsgrad}, and the optimizer parameters are reset after every communication.  We also found it useful to decay the initial learning rate by a factor of $0.99$ after every communication.

\subsection{Parameter Sensitivity Analysis for PFNM}
\label{supp:param_sensitivity}
Our models presented in Section \ref{sec:matching} have three parameters $\sigma^2_0, \gamma_0$ and $\sigma^2=\sigma^2_1=\ldots=\sigma^2_J$. The first parameter, $\sigma^2_0$, is the prior variance of weights of the global neural network. Second parameter, $\gamma_0$, controls discovery of new neurons and correspondingly increasing $\gamma_0$ increases the size of the learned global network. The third parameter, $\sigma^2$, is the variance of the local neural network weights around corresponding global network weights.
We empirically analyze the effect of these parameters on the accuracy for single hidden layer model with $J=25$ batches in Figure \ref{fig:sens}. The heatmap indicates the accuracy on the training data - we see that for all parameter values considered performance doesn't not fluctuate significantly. PFNM appears to be robust to choices of $\sigma_0^2$ and $\gamma_0$, which we set to 10 and 1 respectively in the experiments with single communication round. Parameter $\sigma^2$ has slightly higher impact on the performance and we set it using training data during experiments. To quantify importance of $\sigma^2$ for fixed $\sigma_0^2=10$ and $\gamma_0=1$ we plot average train data accuracies for varying $\sigma^2$ in Figure \ref{fig:J_sigma}. We see that for homogeneous partitioning and one hidden layer $\sigma^2$ has almost no effect on the performance (Fig. \ref{fig:J_sigma_mnist_homo} and Fig. \ref{fig:J_sigma_cifar_homo}). In the case of heterogeneous partitioning (Fig. \ref{fig:J_sigma_mnist_hetero} and Fig. \ref{fig:J_sigma_cifar_hetero}), effect of $\sigma^2$ is more noticeable, however all considered values result in competitive performance.

\end{document}

%% file: math_commands.tex
\usepackage{amsmath,amsfonts,bm}

\def\1{\bm{1}}

\def\rb{{\textnormal{b}}}
\def\rc{{\textnormal{c}}}

\def\rq{{\textnormal{q}}}

\def\rvtheta{{\mathbf{\theta}}}

\def\rvp{{\mathbf{p}}}

\def\rvv{{\mathbf{v}}}

\def\vmu{{\bm{\mu}}}

\def\mB{{\bm{B}}}

\def\mI{{\bm{I}}}

\def\mSigma{{\bm{\Sigma}}}

\DeclareMathAlphabet{\mathsfit}{\encodingdefault}{\sfdefault}{m}{sl}
\SetMathAlphabet{\mathsfit}{bold}{\encodingdefault}{\sfdefault}{bx}{n}

\def\emB{{B}}
\def\emC{{C}}

\DeclareMathOperator*{\argmax}{arg\,max}

%% file: abstract.tex
\begin{abstract}

In federated learning problems, data is scattered across different servers and exchanging or pooling it is often impractical or prohibited. We develop a Bayesian nonparametric framework for federated learning with neural networks. Each data server is assumed to provide local neural network weights, which are modeled through our framework. We then develop an inference approach that allows us to synthesize a more expressive global network without additional supervision, data pooling and with as few as a single communication round. We then demonstrate the efficacy of our approach on federated learning problems simulated from two popular image classification datasets.\footnote{Code is available at \url{https://github.com/IBM/probabilistic-federated-neural-matching}}

\end{abstract}

%% file: intro.tex
\section{Introduction}
The standard machine learning paradigm involves algorithms that learn from centralized data, possibly pooled together from multiple data sources. The computations involved may be done on a single machine or farmed out to a cluster of machines. However, in the real world, data often live in silos and amalgamating them may be prohibitively expensive due to communication costs, time sensitivity, or privacy concerns. Consider, for instance, data recorded from sensors embedded in wearable devices. Such data is inherently private, can be voluminous depending on the sampling rate of the sensors, and may be time sensitive depending on the analysis of interest. Pooling data from many users is technically challenging owing to the severe computational burden of moving large amounts of data, and is fraught with privacy concerns stemming from potential data breaches that may expose a user's protected health information (PHI).

Federated learning addresses these pitfalls by obviating the need for centralized data, instead designing algorithms that learn from sequestered data sources. These algorithms iterate between training local models on each data source and distilling them into a global federated model, all without explicitly combining data from different sources. Typical federated learning algorithms, however, require access to locally stored data for learning.
A more extreme case surfaces when one has access to models pre-trained on local data but not the data itself. Such situations may arise from catastrophic data loss but increasingly also from regulations such as the general data protection regulation (GDPR)~\cite{eu:gdpr}, which place severe restrictions on the storage and sharing of personal data. Learned models that capture only aggregate statistics of the data can typically be disseminated with fewer limitations. A natural question then is, can ``legacy'' models trained independently on data from different sources be combined into an improved federated model? 

Here, we develop and carefully investigate a probabilistic federated learning framework with a particular emphasis on training and aggregating neural network models. We assume that either local data or pre-trained models trained on local data are available. When data is available, we proceed by training local models for each data source, \emph{in parallel}. We then match the estimated local model parameters (groups of weight vectors in the case of neural networks) across data sources to construct a global network.
The matching, to be formally defined later, is governed by the posterior of a Beta-Bernoulli process (BBP)~\citep{thibaux2007hierarchical}, a Bayesian nonparametric (BNP) model that allows the local parameters to either match existing global ones or to create new global parameters if existing ones are poor matches. 

Our construction provides several advantages over existing approaches. First, it decouples the learning of local models from their amalgamation into a global federated model. This decoupling allows us to remain agnostic about the local learning algorithms, which may be adapted as necessary, with each data source potentially even using a different learning algorithm. Moreover, given only pre-trained models, our BBP informed matching procedure is able to combine them into a federated global model without requiring additional data or knowledge of the learning algorithms used to generate the pre-trained models. This is in sharp contrast with existing work on federated learning of neural networks~\cite{mcmahan2017communication}, which require strong assumptions about the local learners, for instance, that they share the same random initialization, and are not applicable for combining pre-trained models. Next, the BNP nature of our model ensures that we recover compressed global models with fewer parameters than the cardinality of the set of all local parameters. Unlike naive ensembles of local models, this allows us to store fewer parameters and perform more efficient inference at test time, requiring only a single forward pass through the compressed model as opposed to $J$ forward passes, once for each local model. While techniques such as knowledge distillation~\citep{hinton2015distilling} allow for the cost of multiple forward passes to be amortized, training the distilled model itself requires access to data pooled across all sources or an auxiliary dataset, luxuries unavailable in our scenario. Finally, even in the traditional federated learning scenario, where local and global models are learned together, we show empirically that our proposed method outperforms existing distributed training and federated learning algorithms~\citep{dean2012large, mcmahan2017communication} while requiring far fewer communications between the local data sources and the global model server.

The remainder of the paper is organized as follows. We briefly introduce the Beta-Bernoulli process in Section~\ref{sec:background} before describing our model for federated learning in Section~\ref{sec:matching}.
We thoroughly evaluate the proposed model and demonstrate its utility empirically in Section~\ref{sec:experiments}. Finally, Section \ref{sec:discussion} discusses current limitations of our work and open questions. 

%% file: background.tex
\section{Background and Related Works}
\label{sec:background}
Our approach builds on tools from Bayesian nonparametrics, in particular the  Beta-Bernoulli Process (BBP)~\citep{thibaux2007hierarchical} and the closely related Indian Buffet Process (IBP)~\citep{griffiths2011indian}. We briefly review these ideas before describing our approach. 

\subsection{Beta-Bernoulli Process (BBP)}

Let $Q$ be a random measure distributed by a Beta process with mass parameter $\gamma_0$ and base measure $H$. That is, $Q | \gamma_0, H \sim \text{BP}(1, \gamma_0 H)$. It follows that $Q$ is a discrete (\emph{not} probability) measure $Q = \sum_i \rq_i \delta_{\rvtheta_i}$ formed by an infinitely countable set of (weight, atom) pairs $(\rq_i,\rvtheta_i) \in [0,1]\times\Omega$. The weights $\{\rq_i\}_{i=1}^\infty$ are distributed by a stick-breaking process \citep{teh2007stick}: $\rc_i \sim \text{Beta}(\gamma_0,1),\, \rq_i = \prod_{j=1}^i  \rc_j$ and the atoms are drawn i.i.d from the normalized base measure $\rvtheta_i \sim H/H(\Omega)$ with domain $\Omega$. In this paper, $\Omega$ is simply $\mathbb{R}^D$ for some $D$. Subsets of atoms in the random measure $Q$ are then selected using a Bernoulli process with a base measure $Q$. That is, each subset $\Tp_j$ with $j = 1, \ldots, J$ is characterized by a Bernoulli process with base measure $Q$, $\Tp_j|Q \sim \text{BeP}(Q)$. Each subset $\Tp_j$ is also a discrete measure formed by pairs $(\rb_{ji},\rvtheta_i) \in \{0,1\}\times\Omega$,
$\Tp_j  := \sum_i \rb_{ji} \delta_{\rvtheta_i}\text{, where } \rb_{ji}|\rq_i \sim \text{Bernoulli}(\rq_i)\,\forall i$ is a binary random variable indicating whether atom $\theta_i$ belongs to subset $\Tp_j$. The collection of such subsets is then said to be distributed by a Beta-Bernoulli process. 

\subsection{Indian Buffet Process (IBP)}
The above subsets are conditionally independent given $Q$. Thus, marginalizing $Q$ will induce dependencies among them. In particular, we have
$\Tp_J | \Tp_1,\ldots,\Tp_{J-1} \sim \text{BeP}\left(H\frac{\gamma_0}{J} + \sum_i\frac{m_i}{J}\delta_{\rvtheta_i}\right)$,
where $m_i = \sum_{j=1}^{J-1} \rb_{ji}$ (dependency on $J$ is suppressed in the notation for simplicity) and is sometimes called the Indian Buffet Process. The IBP can be equivalently described by the following culinary metaphor. Imagine $J$ customers arrive sequentially at a buffet and choose dishes to sample as follows, the first customer tries Poisson$(\gamma_0)$ dishes. Every subsequent $j$-th customer tries each of the previously selected dishes according to their popularity, i.e. dish $i$ with probability $m_i/j$, and then tries Poisson$(\gamma_0/j)$ new dishes. 

The IBP, which specifies a distribution over sparse binary matrices with infinitely many columns, was originally demonstrated for latent factor analysis~\citep{ghahramani2005infinite}.  Several extensions to the IBP (and the equivalent BBP) have been developed, see~\citet{griffiths2011indian} for a review. Our work is related to a recent application of these ideas to distributed topic modeling~\citep{yurochkin2018sddm}, where the authors use the BBP for modeling topics learned from multiple collections of document, and provide an inference scheme based on the Hungarian algorithm~\citep{kuhn1955hungarian}.

\subsection{Federated and Distributed Learning}
Federated learning has garnered interest from the machine learning community of late. \citet{smith2017federated} pose federated learning as a multi-task learning problem, which exploits the convexity and decomposability of the cost function of the underlying support vector machine (SVM) model for distributed learning. This approach however does not extend to the neural network structure considered in our work. \citet{mcmahan2017communication} use strategies based on simple averaging of the local learner weights to learn the federated model. However, as pointed out by the authors, such naive averaging of model parameters can be disastrous for non-convex cost functions. To cope, they have to use a scheme where the local learners are forced to share the same random initialization. In contrast, our proposed framework is naturally immune to such issues since its development assumes nothing specific about how the local models were trained. Moreover, unlike existing work in this area, our framework is non-parametric in nature allowing the federated model to flexibly grow or shrink its complexity (i.e., its size) to account for varying data complexity.

There is also significant work on distributed deep learning~\cite{lian2015asynchronous, lian2017async, moritz2015sparknet, li2014scaling, dean2012large}. However, the emphasis of these works is on scalable training from large data and they typically require frequent communication between the distributed nodes to be effective. Yet others explore distributed optimization with a specific emphasis on communication efficiency~\citep{zhang2013information, shamir2014communication, yang2013trading, ma2015, zhang2015disco}. However, as pointed out by \citet{mcmahan2017communication}, these works primarily focus on settings with convex cost functions and often assume that each distributed data source contains an equal number of data instances. These assumptions, in general, do not hold in our scenario. Finally, neither these distributed learning approaches nor existing federated learning approaches decouple local training from global model aggregation. As a result they are not suitable for combining pre-trained legacy models, a particular problem of interest in this paper.
